\documentclass[10pt, a4paper]{article}

\usepackage{graphicx}% Include figure files
\usepackage{hyperref}
\usepackage{color}
\usepackage[table]{xcolor}% http://ctan.org/pkg/xcolor
\usepackage{multirow}
\usepackage{booktabs}
\usepackage[ruled]{algorithm2e}
\usepackage{subfigure}

\usepackage[english]{babel}
\usepackage{amssymb,amsmath,amsthm, amsfonts}

\RequirePackage[left=2cm,%
                right=2cm,%
                top=2.25cm,%
                bottom=2.25cm,%
                headheight=12pt,%
                letterpaper]{geometry}%

\theoremstyle{definition}

\providecommand{\keywords}[1]{\textbf{\textit{keywords ---}} #1}

\title{Fractal Descriptors of Texture Images Based on the Triangular Prism Dimension}

\author{Jo\~ao B. Florindo$^{1,2}$, Odemir M. Bruno$^{1}$}
\date{}
\begin{document}
\maketitle
\begin{center}
\noindent{$^1$S\~{a}o Carlos Institute of Physics, University of S\~{a}o Paulo, PO Box 369, 13560-970, S\~{a}o Carlos, SP, Brazil.\\Scientific Computing Group - http://scg.ifsc.usp.br}\\
\noindent{$^2$Institute of Mathematics, Statistics and Scientific Computing - University of Campinas\\
	Rua S\'{e}rgio Buarque de Holanda, 651, Cidade Universit\'{a}ria "Zeferino Vaz" - Distr. Bar\~{a}o Geraldo, CEP 13083-859, Campinas, SP, Brasil}
\end{center}

\begin{abstract}
\noindent{This work presents a novel descriptor for texture images based on fractal geometry and its application to image analysis. The descriptors are provided by estimating the triangular prism fractal dimension under different scales \textcolor{black}{with} a weight exponential parameter, followed by \textcolor{black}{dimensionality} reduction \textcolor{black}{using} Karhunen-Lo\`{e}ve transform. The efficiency of the proposed descriptors is tested on two well-known texture data sets, that is, Brodatz and Vistex, both for classification and image retrieval. The novel method is also tested concerning invariances in situations when the textures are rotated or affected by Gaussian noise. The obtained results outperform other classical and state-of-the-art descriptors in the literature and \textcolor{black}{demonstrate} the power of the triangular descriptors in these tasks, suggesting \textcolor{black}{their} use in practical applications of image analysis based on texture features.} 
\end{abstract}
\begin{center}
\end{center}

\section{Introduction}

Fractal theory has presented a growing interest in many applied areas in the last decades, for instance, in Medicine \cite{AKWKMW11,LB09,HL09}, Physics \cite{GGLL12,LLH12,WDX11}, Computer Science \cite{CGW12,GD11,MHW11}, Engineering \cite{WLCWX12,LPY12,XLJLX11}, among many other fields.

Most of these applications employ the fractal dimension to describe objects that \textcolor{black}{should} be classified or simply described in some manner. Fractal dimension provides important information about the object. While in a \textcolor{black}{mathematical} fractal, the dimension measures the ``fractality'', in a real-world object, it expresses the spatial occupation of the structure. More practically, this implies that fractal dimension is capable of capturing important physical and visual attributes, like \textcolor{black}{roughness}, \textcolor{black}{luminance} or the repetition of geometrical patterns.

Despite its importance as a powerful descriptor, fractal dimension is still limited in the representation of more complex structures. This is true mainly in the analysis of real world objects, when the level of ``fractality'' varies along the same object. The literature shows some approaches to solve this problem, \textcolor{black}{such as multifractals} \cite{H01,CI12,BPPV84}, \textcolor{black}{multiscale fractal dimension} \cite{CC00,MCSM02} and \textcolor{black}{fractal descriptors} \cite{BPFC08,FB11,PPFVOB05,FCB10}. This work is focused on the fractal descriptors solution.

Here, we propose a novel fractal descriptor based on the triangular prism fractal dimension \cite{C85} and apply it to the discrimination and retrieval of texture images. In this approach, the fractal dimension is estimated \textcolor{black}{at} different scales of observation and a weight parameter is introduced as an exponent in the sum of the areas, changing the influence of each scale in the final result and ensuring a more complete and flexible description of the image.

The proposed technique is tested over two well-known \textcolor{black}{texture datasets used for benchmark purposes}. The results of the classification and retrieval of such data sets are compared to other classical and recent texture analysis methods. The results confirm that the proposed descriptor is a valuable tool for image analysis tasks.

\color{black}

\section{Related Works}

Texture analysis is a paradigm where the image is described in terms of statistical patterns formed by spatial arrangements of pixel intensities. The first known systematic study on this topic was carried out by Haralick \cite{H67} and his co-occurrence matrices. Since then, a large number of methods on texture analysis have been proposed in the literature. Among the most successful approaches one can mention local binary patterns \cite{PHZA11}, bag-of-features \cite{VZ05}, scale-invariant feature transform \cite{LSP05}, spatial pyramid matching \cite{LSP06}, invariants of scattering transforms \cite{SM13}, fast 
features invariant to rotation and scale of texture \cite{SM15}, and others.

During the last decades, another branch of methods that have presented interesting results in texture analysis, especially on natural images, are those based on fractal geometry, particularly multifractals \cite{XJF09}, multiscale fractal dimension \cite{MCSM02} and fractal descriptors \cite{BPFC08}. In this context, this work proposes the study and application of fractal descriptors based on the estimative of the fractal dimension using a tessellation of triangular prisms \cite{C85}.

Our proposal has some particular characteristics that distinguish it from other approaches in the literature. First, rather than pre-selecting preferable regions in the image as in \cite{LSP05,LSP06}, here all pixels and scales are equally important \textit{a priori}, which simplifies the modelling and interpretation of the texture descriptors. Another difference from methods such as those in \cite{LSP05,SM13,SM15} is that image invariances are not treated explicitly, although the underlying model and multi-scale process ensures that such effect is attenuated in practice. This is confirmed here in the experimental analysis and avoids the use of cumbersome strategies when in many cases invariances are not a critical issue or even when, for example, a rotated texture should be interpreted as a different object. Finally, an important distinction should be done from approaches such as those in \cite{H67,VZ05,PHZA11} where direct relations are established based on te pixel values. Here there is a complete and well-defined physical model behind the statistics extracted from the image, causing it to be more precise in most cases and more robust to deformations usually found in natural structures.

\color{black}

\section{Fractal Theory}
 
Fractal geometry \textcolor{black}{has been applied to} many diverse areas \cite{CGW12,LLH12,HL09,WLCWX12}. This is motivated mainly by the flexibility of fractal theory in \textcolor{black}{modelling natural} objects, which \textcolor{black}{usually} cannot be \textcolor{black}{precisely} represented through conventional Euclidean geometry.

Fractal theory also presents a concise and powerful framework to describe and identify a natural object, based primarily on the fractal dimension concept. Roughly speaking, fractal dimension measures the complexity of a structure. In this case, complexity is related to the property of presenting details at different \textcolor{black}{scales} of observation. In this way, fractal dimension is of particular importance because it is strongly related to fundamental physical and visual features of the object, \textcolor{black}{such as roughness}, \textcolor{black}{luminance}, \textcolor{black}{distribution of colors}, \textcolor{black}{and} others.

The following section describes in a few words some important aspects of fractal geometry theory and its application to texture analysis.

\subsection{Fractal Dimension}

Fractal dimension is formally defined as being the Hausdorff-Besicovitch dimension of a geometrical set of points $X$ that composes the fractal \textcolor{black}{object}.

In order to define the Hausdorff-Besicovitch dimension, \textcolor{black}{initially the Hausdorff measure} $H^s(X)$ should be defined:
\begin{equation}
	H^s(X) = \lim_{\delta \rightarrow 0}\inf \sum_{i=1}^{\infty} |U_{i}|^{s} \mbox{: $\{U_i\}$ is a $\delta$-cover of $X$},
\end{equation}
where \textcolor{black}{$|U_i|$ states for the diameter of $U_i$ and $\{U_i\}$ is a $\delta$-cover of $X$ iff $X \subset \cup_{i=1}^{\infty}U_i$ with $|U_i| \leq \delta$, for all $i$.}

\textcolor{black}{Analyzing} the behavior of $H^S$ against $s$, \textcolor{black}{one} observes that $H^s$ jumps from $\infty$ to $0$ at a particular real non-negative value of $s$. This value is the \textcolor{black}{Hausdorff-Besicovitch} or fractal dimension \textcolor{black}{of $X$}.

Although the above definition is consistent and \textcolor{black}{can} be applied to any set of points immersed in the Euclidean space, it \textcolor{black}{shows} to be difficult or even impracticable in many situations where the fractal dimension of an object has to be estimated. With the aim of simplifying the computation in such situations, an approximate discrete version of the \textcolor{black}{Hausdorff-Besicovitch} dimension, known as similarity dimension $D_s$, can be defined by:
\textcolor{black}{
\begin{equation}\label{eq:FD}
	D_s = -\lim_{\epsilon \rightarrow 0}\frac{\log(N)}{\log(\epsilon)},
\end{equation}
}
where $N$ is the number of \textcolor{black}{``rulers''} with length $\epsilon$ used to cover the fractal object. Actually, $N$ is a metric that can be generalized to a large sort of measures, both in spatial and frequency domain. This gives rise to a lot of methods for estimation of fractal dimension \cite{F86}, like Bouligand-Minkowski, box-counting, Fourier, etc. The triangular prism dimension employed in this work is an example of method derived from the similarity dimension.

\subsection{Triangular Dimension}\label{sec:tri}

This method for the estimation of \textcolor{black}{the} fractal dimension \textcolor{black}{of objects represented in a gray-level image}, proposed in \cite{C85}, is based on the relation between the surface area of a triangular tessellation of the gray-level map and the dimension of the base of each triangle.

The image is divided into a grid of squares with side-length $\epsilon$. For each square, a triangular prism is constructed using the pixel intensities in each corner as the heights and a central point whose height is given by the average of the corner heights. Thus in a gray-level image $I$, \textcolor{black}{let $a$, $b$, $c$ and $d$} be the pixel intensities delimiting the grid square, such that:
\textcolor{black}{
\begin{equation}
	a = I(i,j); b = I(i+\epsilon,j);c = I(i+\epsilon,j+\epsilon); d = I(i,j+\epsilon).
\end{equation}
}
The center of the prism has height $e$ given by the simple average:
\begin{equation}
	e = (a+b+c+d)/4.
\end{equation}

\begin{figure}[!htpb]
\centering
\includegraphics[width=0.8\textwidth]{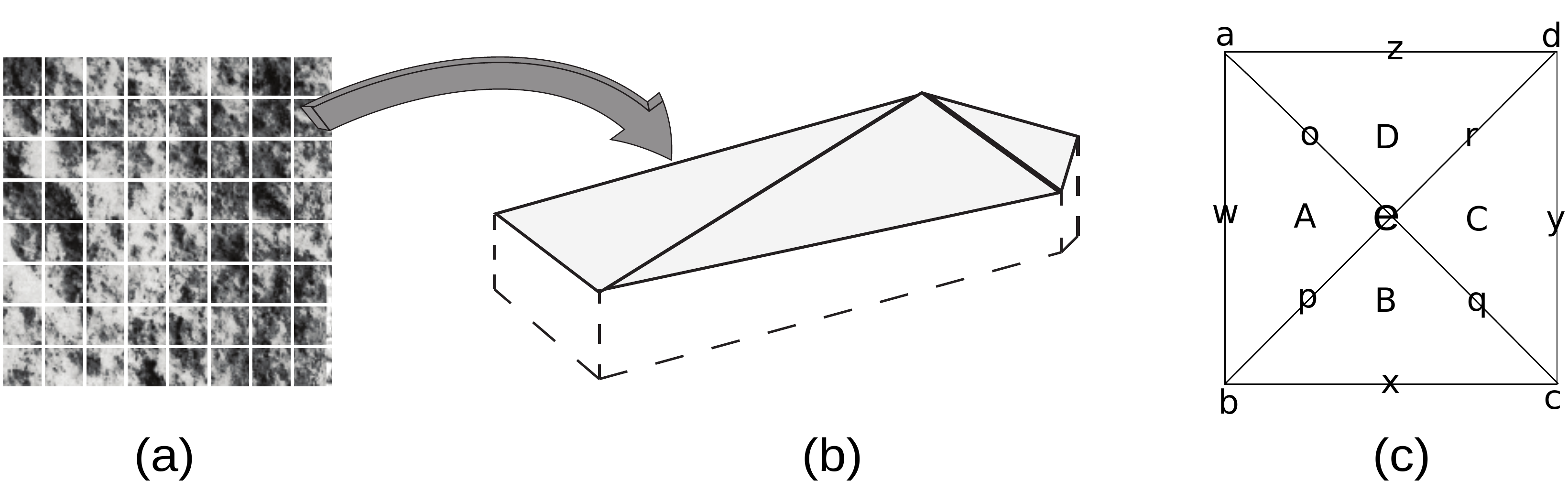}
\caption{Triangular prisms. (a) Gray-level image. (b) A sample prism. (c) Projected prism.}
\label{fig:tri}
\end{figure}

Figure \ref{fig:tri} depicts a scheme of the prism construction.

The set of prisms composes a tessellation surface, such that the total area of this surface can be computed by using some geometrical procedures \textcolor{black}{(Heron formulas)}. Thus for each prism, the \textcolor{black}{semi-perimeter} of each face $A$, $B$, $C$ and $D$ is given by:
\textcolor{black}{
\begin{equation}
	s_a = 1/2(w+p+o);s_b = 1/2(x+p+q);s_c = 1/2(y+q+r);s_d = 1/2(z+o+r).
\end{equation}
where $w$, $x$, $y$, $z$, $o$, $p$, $q$ and $r$ are the segments as labelled in Figure \ref{fig:tri} (c)}:
\begin{equation}
\begin{array}{cc}

	w = \sqrt{(b-a)^2+\epsilon^2};x = \sqrt{(c-b)^2+\epsilon^2};\\
	y = \sqrt{(d-c)^2+\epsilon^2};z = \sqrt{(a-d)^2+\epsilon^2}\\
	o = \sqrt{(a-e)^2+(\frac{\sqrt{2}}{2}\epsilon)^2};p = \sqrt{(b-e)^2+(\frac{\sqrt{2}}{2}\epsilon)^2};\\
	q = \sqrt{(c-e)^2+(\frac{\sqrt{2}}{2}\epsilon)^2};r = \sqrt{(d-e)^2+(\frac{\sqrt{2}}{2}\epsilon)^2}.
\end{array}
\end{equation}
The total area of each triangular prism is provided by \textcolor{black}{the sum of the areas of the four faces}:
\begin{equation}\label{eq:area}
	S_{ij,\epsilon} = A+B+C+D.
\end{equation}
where \textcolor{black}{the area of each face is given by:
\begin{equation}
\begin{array}{cc}
	A = \sqrt{s_a(s_a-w)(s_a-p)(s_a-o)};B = \sqrt{s_b(s_b-x)(s_b-p)(s_b-q)};\\
	C = \sqrt{s_c(s_c-y)(s_c-q)(s_c-r)};D = \sqrt{s_d(s_d-z)(s_d-o)(s_d-r)}.
\end{array}	
\end{equation}}
The total area of the surface is computed by summing the area of each prism in the grid with step $\epsilon$:
\begin{equation}
	S(\epsilon) = \sum_{i',j' \in \mathfrak{G}_{\epsilon}}{S_{i'j',\epsilon}},
\end{equation}
where $\mathfrak{G}_{\epsilon}$ is the set of points in the grid with step $\epsilon$.

This procedure is repeated for a range of values of $\epsilon$ and in each step the area $S(\epsilon)$ is estimated. The fractal dimension $D$ is extracted from the log-log relation between $\epsilon$ and $S$, such that, $D = 2 - \alpha$, where $\alpha$ is the slope of a straight line fit to the curve $\log(\epsilon) \times \log(S(\epsilon))$.

\section{Fractal Descriptors}

Fractal descriptors are an extension of fractal dimension concept. Actually, although fractal dimension is \textcolor{black}{an important descriptor} it is still insufficient to represent \textcolor{black}{more complex systems}. We \textcolor{black}{can easily observe distinct} fractals \textcolor{black}{with} the same fractal dimension despite their completely different \textcolor{black}{appearance}. Such situation is \textcolor{black}{even more complicated} when we deal with objects from the real world, which are not real fractals. In these structures, we find different levels of ``fractality'' according to the \textcolor{black}{observed scale} or even to the spatial region analyzed. In this context we need a tool capable of \textcolor{black}{modelling} the object in all its extension.

\begin{figure}[!htpb]
\centering
\includegraphics[width=0.6\textwidth]{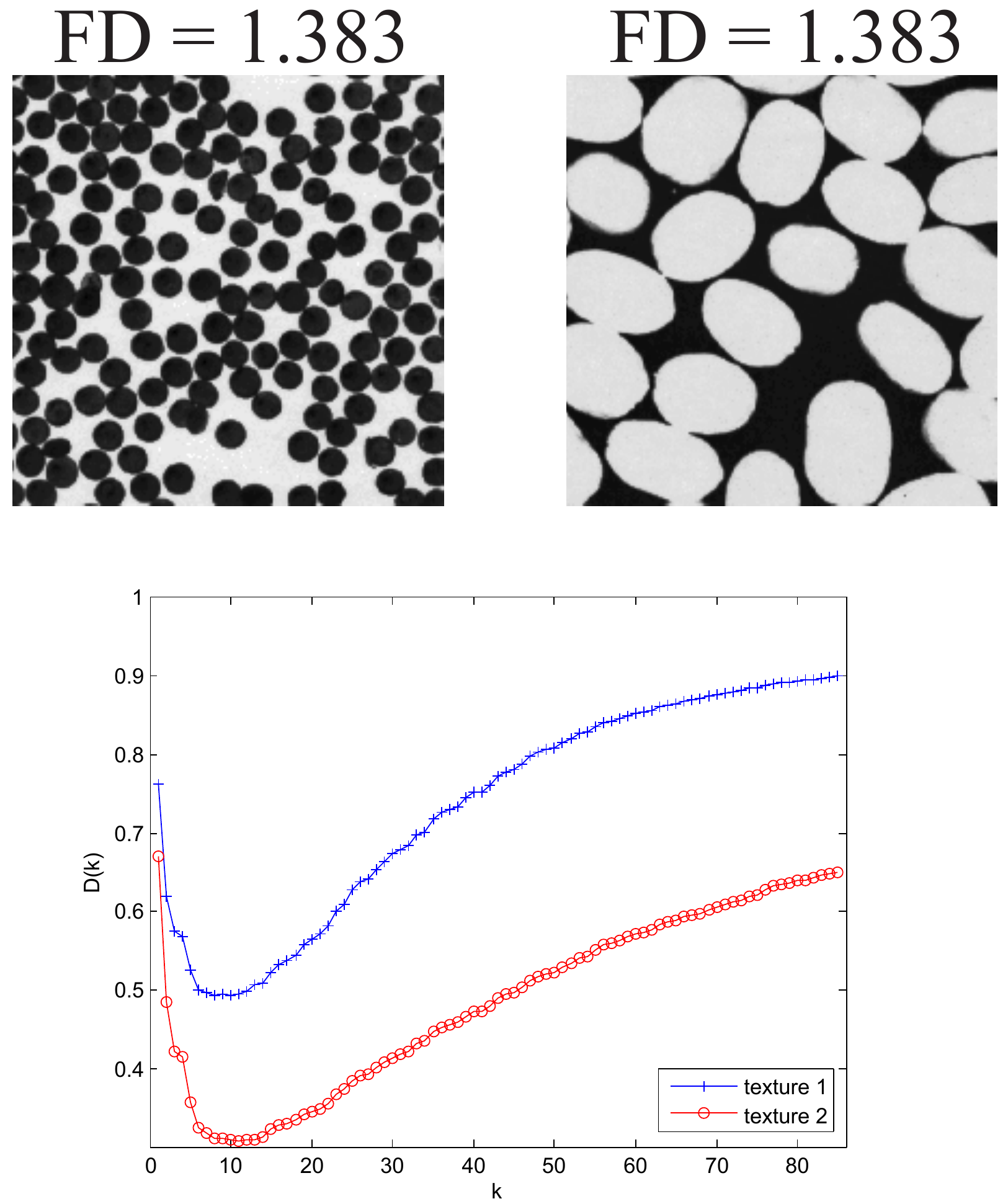}
\caption{\textcolor{black}{On top}, two distinct textures with similar fractal dimension (in this case, estimated through Bouligand-Minkowski approach). At bottom, \textcolor{black}{curves of fractal descriptors} corresponding to each image and \textcolor{black}{illustrating their discrimination power}.}
\label{fig:desc}
\end{figure}

Fractal descriptors \cite{BPFC08,FB11,PPFVOB05,FCB10} \textcolor{black}{constitute} a solution to fill this gap, \textcolor{black}{making} possible a richer analysis of \textcolor{black}{fractal characteristics present in the object}. Figure \ref{fig:desc} illustrates two distinct texture images whose fractal dimensions (FD) are similar. \textcolor{black}{Hence using only} the FD estimation is not enough \textcolor{black}{to} distinguish the images. For the same images, Figure \ref{fig:desc} shows the normalized fractal descriptor curves, which demonstrates \textcolor{black}{to be} capable of discriminating the textures in a straightforward manner. 

Essentially, the purpose of fractal descriptors is to estimate the fractal dimension under different scales, providing information of different patterns and arrangements present in the structure. A natural candidate to allow this analysis is the power-law \textcolor{black}{relation} intrinsic to the fractal dimension $D$. From the \textcolor{black}{similarity dimension} we can write:
\textcolor{black}{
\begin{equation}\label{eq:FD2}
	D = -\lim_{\epsilon \rightarrow 0}\frac{\log(\mathfrak{M}(\epsilon))}{\log(\epsilon)},
\end{equation} 
}
where $\mathfrak{M}$ is any measure, related to the spatial or frequency distribution of the object and $\epsilon$ is the scale parameter. The power-law relation may be stated in a quite simple fashion:
\begin{equation}
	\mathfrak{M} \propto \epsilon^{-D}.
\end{equation}

Fractal descriptors consist in extracting features from the following function:
\begin{equation}\label{eq:multi}
	u:\log(\epsilon) \rightarrow \log(\mathfrak{M}).
\end{equation}

The function $u$ may be used in different manners \textcolor{black}{either} directly, as in \cite{BCB09}, or after a multiscale transform \cite{BPFC08} or by the application of Principal Component Analysis or still Functional Data Analysis \textcolor{black}{as} in \cite{FCB10}, among many other possibilities.

\section{Proposed Method}

This work proposes a novel fractal descriptor based on the fractal dimension estimated by triangular prisms. In this case, the fractality function $\mathfrak{M}(\epsilon)$ corresponds to the area function $S(\epsilon)$. Following the general idea of fractal descriptors, the proposed features are provided by \textcolor{black}{$\log(S(\epsilon))$}. 

\textcolor{black}{Furthermore}, to improve the ability of identifying multiscale patterns along the texture, the proposed method introduces an exponent weight $\alpha$ to the area sum $S(\epsilon)$, which now has the following expression:
\begin{equation}\label{eq:desc}
	S^{\alpha}(\epsilon) = \sum_{i',j' \in \mathfrak{G}_{\epsilon}}{S_{i'j',\epsilon}^{\alpha}},
\end{equation}
where $S_{i'j',\epsilon}$ is defined as in Equation \ref{eq:area}.

For image analysis purposes, a range of values $\alpha$ is chosen empirically and the following values are taken into account to compose the descriptors $D$:
\begin{equation}
	D = \bigcup_{1 \leq \epsilon \leq \epsilon_{max},\alpha_{min} \leq \alpha\leq \alpha_{max}}S^{\alpha}(\epsilon).
\end{equation}

Finally, since the above expression generates a too much large set of features, a reduction of dimensionality is necessary. Such procedure is carried out by a Karhunen-Lo\`{e}ve (KL) transform. Thus let $M_D$ be the feature matrix, that is, a matrix \textcolor{black}{where} each line corresponds to the vector $D$ of a sample and each column is a descriptor. The covariance matrix $M_C$ is defined by:
\begin{equation}
	M_C(i,j) = \frac{\sum_{i=1}^{n}{(M(.,i)-\overline{M(.,i)})(M(.,j)-\overline{M(.,j)})}}{n-1},
\end{equation}
where $n$ is the number of descriptors in $D$, $M(i,.)$ is the $i^{th}$ column of $M$ and $\overline{(.)}$ is the average of the vector. In the following a second matrix $U$ is defined as having in each column the eigenvectors of $M_C$, sorted according to decreasing values of eigenvalues. Finally, the KL transform $\tilde{M}_D$ of $M_D$ is obtained by
\begin{equation}
	\tilde{M}_D = U^T M_D.
\end{equation}
Now, $\tilde{M}_D$ is the new matrix of features. It has the same dimensions of $M_D$ and each line in $\tilde{M}_D$ contains the final descriptors of the respective sample. Based on the KL transform theory, the first descriptors are the most meaningful for the analysis. Here, the descriptors are considered in this order and, as it will be described in Results section, the number of such descriptors is varied between 1 and a predefined maximum, to find out the best configuration empirically.

\begin{figure}[!htpb]
\centering
\includegraphics[width=\textwidth]{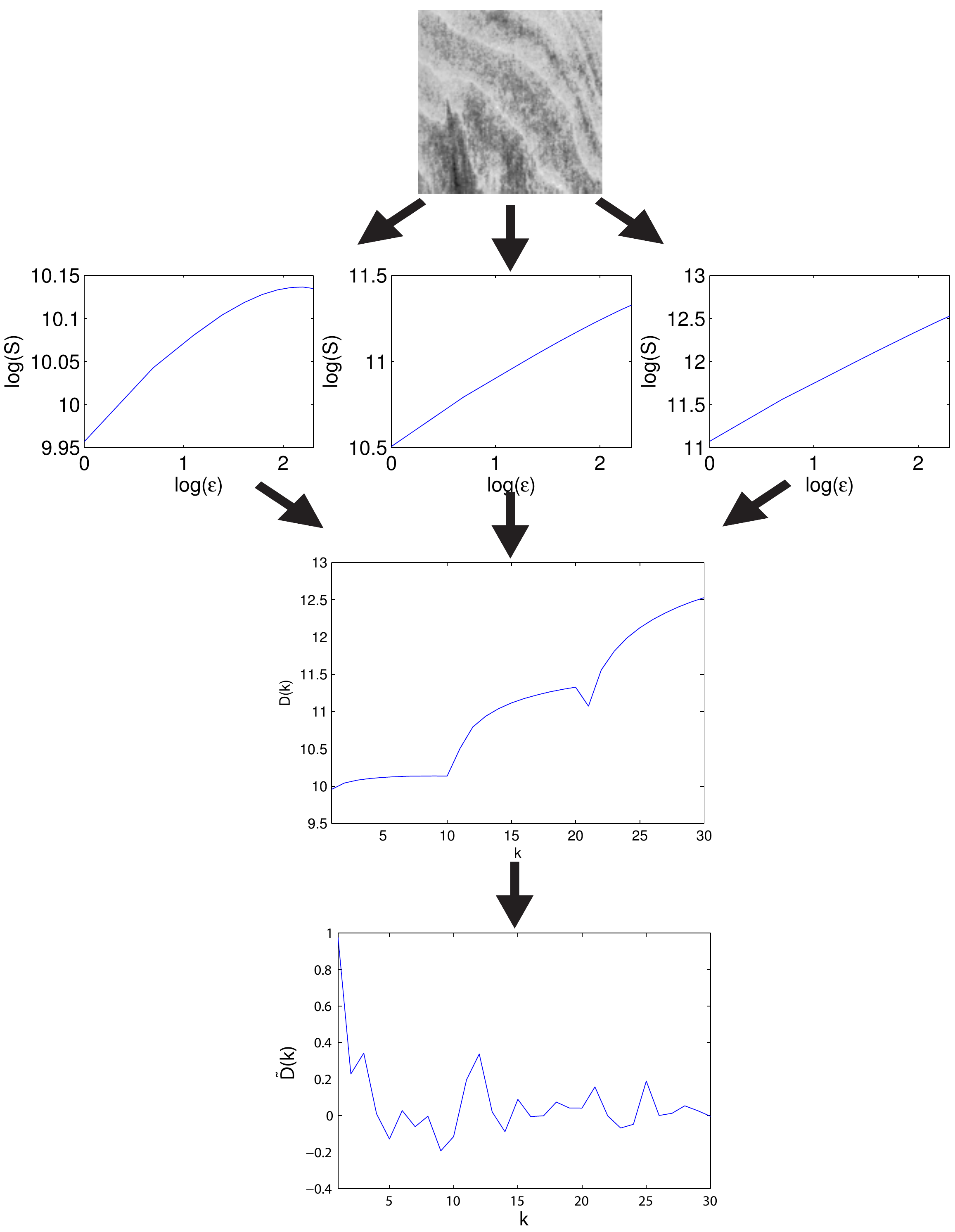}
\caption{A diagram illustratting the composition of the proposed descriptors. From top to bottom, the texture analyzed, the log-log curves of triangular dimension with three exponents (0.1, 0.3 and 0.5), the concatenated descriptors and the final descriptors after the KL transform.}
\label{fig:method}
\end{figure}

Figure \ref{fig:method} summarizes the described steps in a visual diagram.

\begin{figure}[!htpb]
\centering
\includegraphics[width=0.8\textwidth]{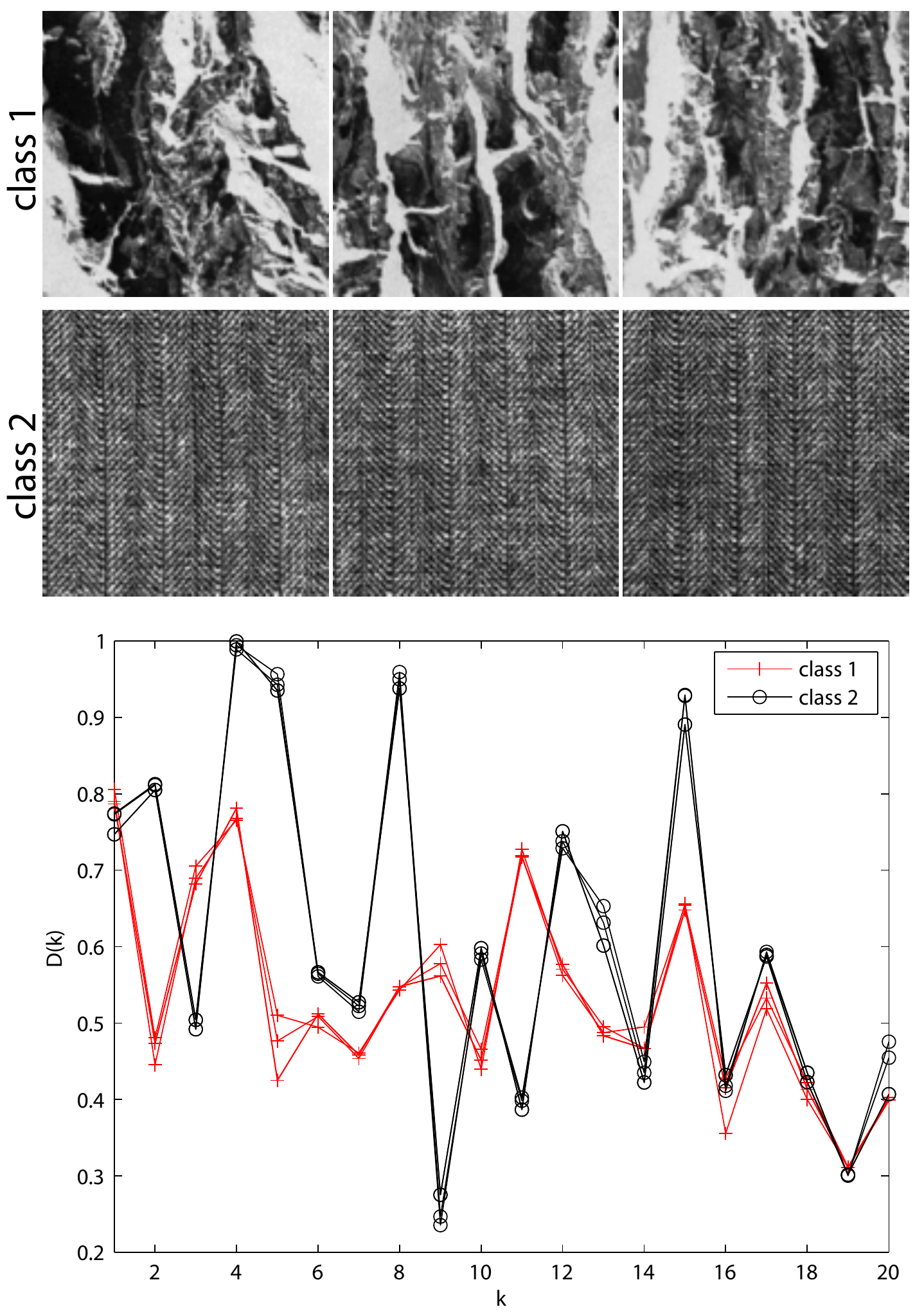}
\caption{A graphical representation of the proposed descriptors for two classes of textures. Even visually, the fractal features \textcolor{black}{discriminate} the images.}
\label{fig:descriptor}
\end{figure}

The proposed method combines two types of information to provide a rich and reliable image descriptor. The first is the complexity, measured by the fractal dimension. Such property is strongly correlated with physical characteristics of the object. The second one is the multiscale analysis that is accomplished by the parameter $\epsilon$. Using the curve of \textcolor{black}{$S(\epsilon)$} ensures that the complexity information is expressed along a range of scales. Finally, the exponent $\alpha$ provides an empirical weight for each scale, making the proposed descriptors more flexible to address the description of so diverse objects. All this combination yields a method capable of identifying and discriminating objects even in severe situations as when there is high variability among elements of a same class and/or high similarity among samples of different classes. Figure \ref{fig:descriptor} visually illustrates the discrimination of two classes of textures by the proposed method.

\color{black}

\subsection{Motivation}

\subsubsection{Fractal Modelling}

To better understand how and why tringular prisms work on the image, we should look at its mathematical interpretation. The classical analysis of the relation between triangular prisms and fractal theory is that adopted in \cite{C85}. It relies on the idea of extending the ``walking-dividers'' approach \cite{M82} to a two-dimensional manifold. %Although such rationale is valid, it does not allow a more rigorous formalism to justify its fractal essence. 
Here we propose a second interpretation based on the theory of fractional Brownian motion (fBm) \cite{F86}. 

Our first objective is to establish the relation between the prism areas and the pixel gray levels in the image. We illustrate the computation in face $A$. As $e$ can be written as a function of $(a,b,c,d)$ we can rewrite $o$ and $p$:
\begin{equation}\label{eq:op}
	\begin{array}{cc}
		o = \sqrt{\left( \frac{3a-b-c-d}{4} \right)^2 + \frac{1}{2}\epsilon^2}
		&
		p = \sqrt{\left( \frac{3b-a-c-d}{4} \right)^2 + \frac{1}{2}\epsilon^2}
	\end{array}
\end{equation}
We can also plug the definition of $s_a$ into the area $A$:
\begin{equation}
\begin{array}{l}
A = \sqrt{\left( \frac{w+o+p}{2} \right)\left( \frac{-w+o+p}{2} \right)\left( \frac{w+o-p}{2} \right)\left( \frac{w-o+p}{2} \right)} = \\
= \frac{\sqrt{-o^4 - p^4 - w^4 + 2o^2p^2 + 2o^2w^2 + 2p^2w^2}}{4} = \\
= \frac{\sqrt{4p^2w^2 - (o^2 - p^2 - w^2)^2}}{4}.
\end{array}		
\end{equation}
Replacing $w$, $p$ and $o$ with their respective representation in terms of $a$, $b$, $c$ and $d$ we end up with the following expression:
\footnotesize
\begin{equation}
	\begin{array}{l}
		A = \frac{
		\sqrt{4(\left( \frac{3b-a-c-d}{4} \right)^2 + \frac{1}{2}\epsilon^2)((b-a)^2 + \epsilon^2) - \left[ (\left( \frac{3a-b-c-d}{4} \right)^2 + \frac{1}{2}\epsilon^2) - (\left( \frac{3b-a-c-d}{4} \right)^2 + \frac{1}{2}\epsilon^2) - ((b-a)^2 + \epsilon^2) \right]^2}
		}{4}\\	
		= \frac{\epsilon}{4} \sqrt{1.25a^2 + 1.25b^2 + 0.25c^2 + 0.25d^2 - 1.5ab - 0.5ac - 0.5ad - 0.5bc - 0.5bd + 0.5cd + \epsilon^2}\\
		= \frac{\epsilon}{8} \sqrt{5a^2 + 5b^2 + c^2 + d^2 - 6ab - 2ac - 2ad - 2bc - 2bd + 2cd + 4\epsilon^2}\\		
	\end{array}		
\end{equation} 
\normalsize
Rearranging terms we can write the above expression as
\begin{equation}\label{eq:TrifBm}
	\begin{array}{l}
		A = \frac{\epsilon}{4}\sqrt{\left( \frac{a+b+c+d}{2} \right)^2 + (a-b)^2 - (a+b)(c+d) + \epsilon^2}\\
		=  \frac{\epsilon}{4}\sqrt{\frac{((a+b)-(c+d))^2}{4} + (a-b)^2 + \epsilon^2}\\
		= \frac{\epsilon}{8}\sqrt{4(a-b)^2 + ((a+b)-(c+d))^2 + 4\epsilon^2}\\
	\end{array}		
\end{equation}
The expression enclosed by the square root contains three squared terms and the leading one is $(a-b)^2$. The points in the image where the intensities are $a$ and $b$ are separated by a distance $\epsilon$. Therefore the distribution of this term is classically related to the fBm process \cite{F86}.

An fBm process $B_H(t)$ is a Gaussian non-stationary stochastic process, with mean zero and variance $\sigma$, whose covariance statistics satisfies
\begin{equation}
	E(B_H(t)B_H(s)) = \frac{\sigma^2}{2}(|t|^{2H} + |s|^{2H} - |t-s|^{2H}),
\end{equation}
where $E$ is the expected value and $H$ is a parameter named Hurst exponent. Another way of defining the same process is to write
\begin{equation}\label{eq:VAR}
	E((B_H(t_1)-B_H(t_2))^2) \propto |t_1 - t_2|^{2H}.
\end{equation}

Seminal works on fractal geometry in images, such as those of Mandelbrot \cite{M82} and Pentland \cite{P84}, already demonstrated that fractal characteristics in images reflect fractality in the physical process originating the pictured object. Pentland even carried out a survey to discover that such relation can be perceptually confirmed by human beings. Since then, a number of studies have been based on this assumption and fBm has been considered the canonical representation for such physical processes giving rise to fractal images. 

In this context, the expression (\ref{eq:VAR}) has been paramount. If the image obeys the statistics in (\ref{eq:VAR}) with an acceptable $p$-value the image can be analyzed as a statistical fractal. Furthermore, as demonstrated in \cite{F86}, the fBm has fractal dimension (in the sense of Hausdorff-Besicovitch) equals to $3-H$ with probability $1$.

In practice it is usual to verify the statistics (\ref{eq:VAR}) in an image $I$ by checking the curve of $\ln \epsilon$ $vs.$ $\ln <(I(x)-I(y))^2>$ where $<.>$ stands for the average and $x$ and $y$ are points separated by $\epsilon$. This is in essence what is expressed by the term $(a-b)^2$ in (\ref{eq:TrifBm}). The second squared term ($((a+b)-(c+d))^2$) can be seen as a correction taking into account the influence of the neighbor pixels in the fBm covariance. Similarly, the parameter $\epsilon$ appears as a weight expressing the scale size.

We can conclude from this that the triangular prisms are capable of representing two classical viewpoints in fractal geometry, to know, a geometrical interpretation such as ``walking-dividers'' and a statistical approach derived from fBm. Back to the context of fractal descriptors, the statistics (\ref{eq:VAR}) does not to fit perfectly the image. Rather what we expect to find is how the image is more or less close to a fractal in each scale. And this analysis highly depends on the richness of the measure provided by the method employed to estimate the fractal dimension. In this way, a method with a more complete description of the fractal process, both in a geometrical and a statistical sense, tends to be more appropriate than other classical methods in the literature, hence explaining the success achieved by the proposed approach.

\subsubsection{Role of exponent $\alpha$}

Now we turn our attention to the exponent $\alpha$ in (\ref{eq:desc}). Although a generalized version of the multinomial theorem with real exponents exists, it is not straightforward to be interpreted. Therefore we opted for employing MacLaurin series up to the second order. The general expression for a real multi-variable function is:
\begin{equation}
	\begin{array}{l}
		f(x_1,x_2,\dots,x_n) \approx f(0,0,\dots,0) + \sum_{i=1}^{n}\frac{\partial f}{\partial x_i}(0,0,\dots,0)x_i + \\ + \frac{1}{2!}\sum_{i=1}^{n}\sum_{j=1}^{n}\frac{\partial f}{\partial x_i \partial x_j}(0,0,\dots,0)x_ix_j +\\ 
		+ \frac{1}{3!}\sum_{i=1}^{n}\sum_{j=1}^{n}\sum_{k=1}^{n}\frac{\partial f}{\partial x_i \partial x_j \partial x_k}(0,0,\dots,0)x_ix_jx_k + \dots		
	\end{array}		
\end{equation}
We will replace the first term in the summation (\ref{eq:desc}) by $(1+x_1)$ and all the others by $x_2,x_3,x_4,\dots$ resulting in the function $(1+x_1+x_2+\dots+x_n)^\alpha$. Plugging the power function derivatives:
\begin{equation}
	\begin{array}{l}
		(1+x_1+x_2+\dots+x_n)^\alpha \approx 1 + (\alpha-1)\sum_{i=1}^{n}x_i + \frac{(\alpha-1)^2}{2} \sum_{i=1}^{n}\sum_{j=1}^{n}x_ix_j +\\
		+ \frac{(\alpha-1)^3}{6}\sum_{i=1}^{n}\sum_{j=1}^{n}\sum_{k=1}^{n}x_ix_jx_k + \dots
	\end{array}		
\end{equation}
Whereas the first summation term is related to the mean of $x_i$ the following ones relate to the correlation between groups of $2,3,4,\dots$ variables. The exponent $\alpha$ gives weight to these terms and the higher the exponent the larger the influence of the correlation within larger groups of variables. Correlation is a well-established statistics in texture images as it is related to Haralick second-order statistics \cite{H67} and $\alpha$ quantify such property in terms of the prism areas and indirectly on the local fractality of the image.

Altogether the combination of an accurate fractal analysis with an intrinsic correlation is what allows for the analysis proposed here to be powerful even in the most challenging scenarios such as when we have high variability among images of a same group or when images from different groups can be confused by more traditional approaches in the literature, where the modelling may be not sufficiently solid and flexible at the same time neither metrics employed can express the parameters of the underlying model with the necessary accuracy.

\section{Algorithm}

The following pseudo-code contains the algorithm for the triangular prism descriptors. Essentially it is not too much than a mere translation of the mathematical expressions in Section \ref{sec:tri}. A simplified version written in Matlab$^{\copyright}$/Octave can also be found at \footnote{www....};

	\begin{algorithm}[H]
		\caption{Triangular prism image descriptors}
		%\SetKwInOut{Input}{input}\SetKwInOut{Output}{output}		
		\KwIn{Image $I$ with dimension $M \times N$; Exponent $\alpha$}
		\KwOut{Descriptors $Desc$}
		$k \leftarrow 1; \delta \leftarrow 1$ \;
		\While{$\delta \leq \log_2(\min(M,N))$}{
			$\epsilon \leftarrow \lfloor 2^\delta \rfloor$\;
			\For{$i=1$ \KwTo $M$}{
				\For{$j=1$ \KwTo $N$}{
					$a \leftarrow I(i,j); b \leftarrow I(i+\epsilon,j);c \leftarrow I(i+\epsilon,j+\epsilon); d \leftarrow I(i,j+\epsilon)$\;
					$e \leftarrow (a+b+c+d)/4$\;
					$w \leftarrow \sqrt{(b-a)^2+\epsilon^2}; x \leftarrow \sqrt{(c-b)^2+\epsilon^2}$ \;
					$y \leftarrow \sqrt{(d-c)^2+\epsilon^2}; z \leftarrow \sqrt{(a-d)^2+\epsilon^2}$ \;
					$o \leftarrow \sqrt{(a-e)^2+(\frac{\sqrt{2}}{2}\epsilon)^2}; p \leftarrow \sqrt{(b-e)^2+(\frac{\sqrt{2}}{2}\epsilon)^2}$ \;
					$q \leftarrow \sqrt{(c-e)^2+(\frac{\sqrt{2}}{2}\epsilon)^2}; r \leftarrow \sqrt{(d-e)^2+(\frac{\sqrt{2}}{2}\epsilon)^2}$ \;
					$s_a \leftarrow 1/2(w+p+o);s_b = 1/2(x+p+q)$ \;
					$s_c \leftarrow 1/2(y+q+r);s_d = 1/2(z+o+r)$ \;
					$A \leftarrow \sqrt{s_a(s_a-w)(s_a-p)(s_a-o)}$ \;
					$B \leftarrow \sqrt{s_b(s_b-x)(s_b-p)(s_b-q)}$ \;
					$C \leftarrow \sqrt{s_c(s_c-y)(s_c-q)(s_c-r)}$ \;
					$D \leftarrow \sqrt{s_d(s_d-z)(s_d-o)(s_d-r)}$ \;
					$Desc(k) \leftarrow Desc(k) + (A+B+C+D)^\alpha$ \;
				}
			}	
			$k \leftarrow k+1$ \;			
			$\delta \leftarrow \delta+1$ \;
		}
		\Return{x}\;
	\end{algorithm}
	
\color{black}

\section{Experiments}

The proposed method is tested over two data sets, that is, Brodatz \cite{B66} and Vistex \cite{Vistex}. Brodatz is a gray-level texture database composed by 111 texture images divided each one into 16 windows with 128$\times$128 pixels without overlapping. Vistex is composed by 54 color texture images, each one divided into 16 128$\times$128 windows. The color images are converted into gray-level ones before the descriptor analysis. In both experiments, each entire image is a class and each window corresponds to a sample.

The classification is performed by a Linear Discriminant Analysis (LDA) method \cite{DH00}. The classification of each \textcolor{black}{data set} is achieved by inputting the proposed descriptors and other well-known texture descriptors to the classifier, using a 10-fold scheme, where one tenth part of the database is used for testing and the remaining samples for training. The classification results are compared with other state-of-the-art and classical texture descriptors in the literature, that is, Local Binary Patterns (LBP) \cite{PHZA11}, Gabor-wavelets \cite{MM96}, Multifractal spectrum \cite{XJF09}, Gray Level Co-occurrence Matrix (GLCM) \cite{H67} and Fourier descriptors \cite{GW02}.

\section{Results}

\subsection{Parameter Settings}

\begin{figure}[!htpb]
		\centering
		\begin{tabular}{cc}
	    \includegraphics[width=0.45\textwidth]{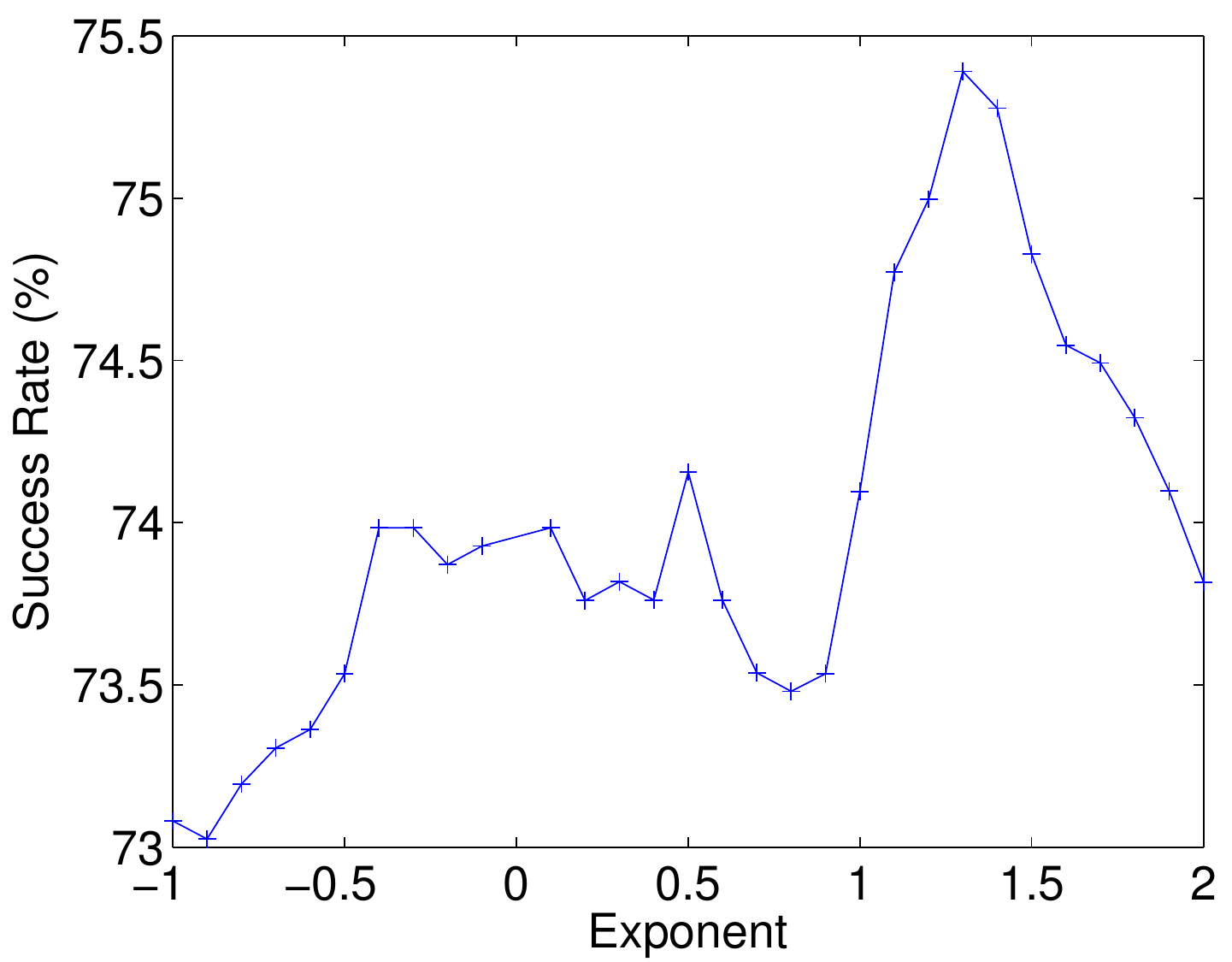} &
		\includegraphics[width=0.45\textwidth]{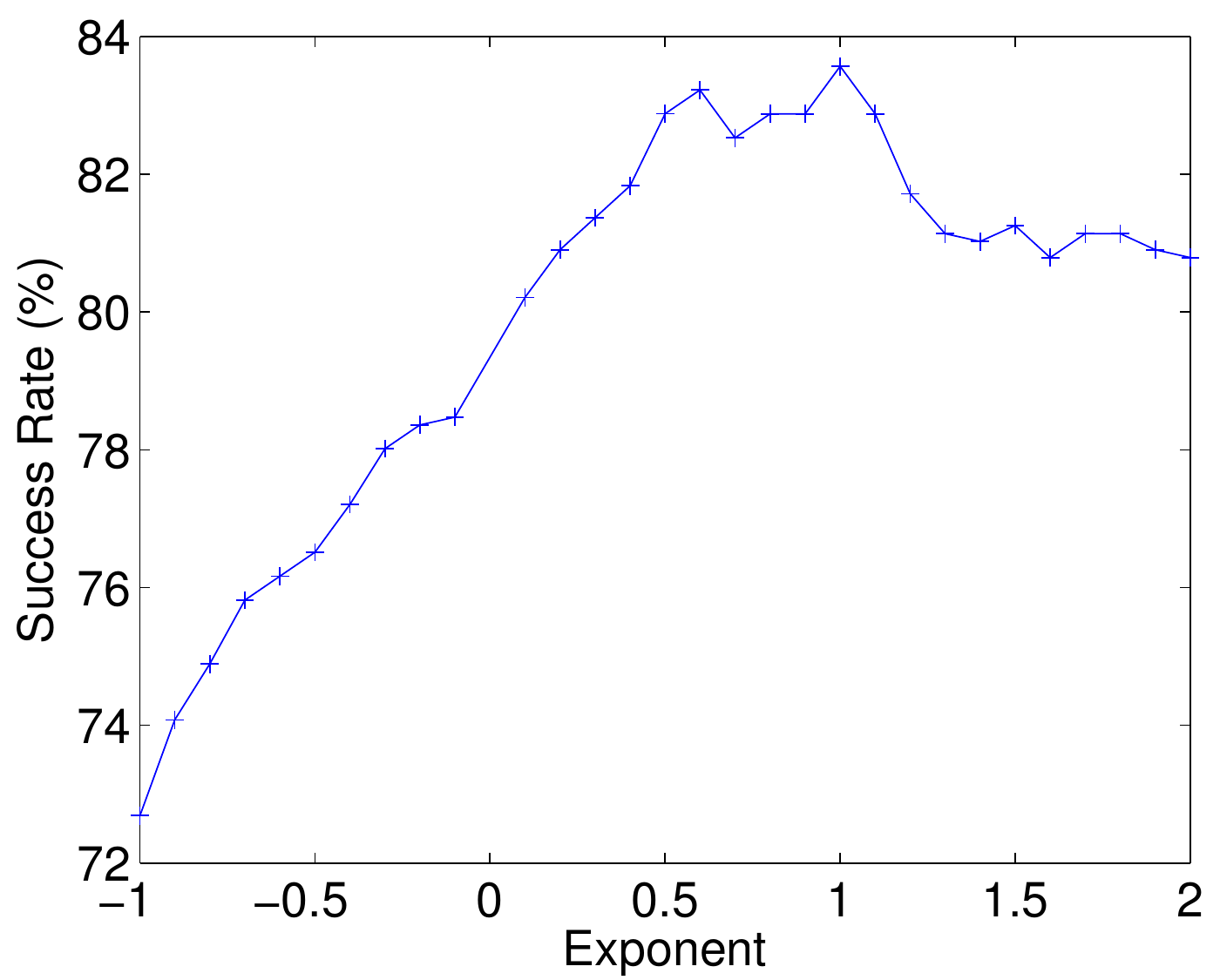}\\
		(a) & (b)\\
		\end{tabular}
	 	\caption{Success rate when using different values for the exponent of the triangular areas in: (a) Brodatz database and (b) Vistex database.}
           \label{fig:exponent}                                  
   \end{figure}

The proposed method depends on an exponential parameter that can be \textcolor{black}{tuned} to find the most suitable descriptors for each type of image. In this way, to find the best parameters is a fundamental task to ensure a robust solution. In real situations, these variables can be empirically evaluated over the training set. Figure \ref{fig:exponent}(a) shows the precision in the classification of Brodatz data set when using exponents ranging between -1 and 2. Outside this range there is no significant gain. Based on this plot, the triangular descriptors with exponents between 1 and 1.8 were concatenated and summarized through PCA to provide the texture descriptors used \textcolor{black}{in} the following experiments. \textcolor{black}{Ten} descriptors are used for each exponent.

A similar test was accomplished over Vistex database giving rise to the curve in Figure \ref{fig:exponent} (b). In this case, exponents with values between 0.4 and 1.2 were selected to compose the proposed descriptors.

\subsection{Image Retrieval}

Now the proposed method is applied to practical problems in image analysis and its performance is compared to other texture descriptors proposed in the literature. The first task discussed is \textcolor{black}{image} retrieval, when the user \textcolor{black}{inputs} a sample and \textcolor{black}{tries} to recall similar images (from the same class). Each attempt to recall the expected sample is named \textcolor{black}{``query''} and the aim is to obtain the maximum ratio of expected samples with the minimum possible of queries. A complete description of this trade-off is given by the precision/recall curve. Precision is the ratio between the number of expected samples returned and the number of queries. Recall is the ratio between the number of right guesses and the total number of expected samples in the database.

\begin{figure}[!htpb]
\centering
\includegraphics[width=.6\textwidth]{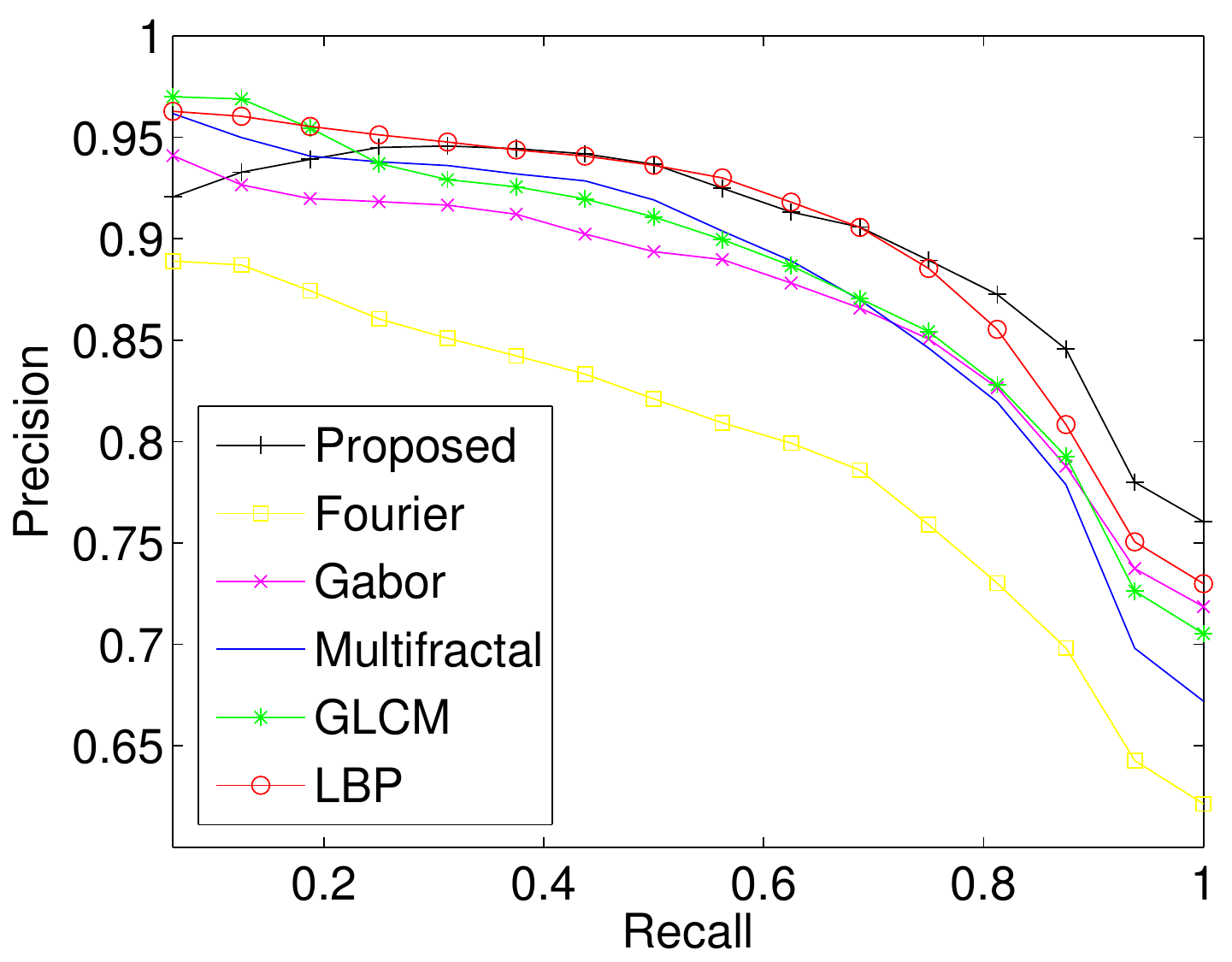}
\caption{Precision/Recall curves for Brodatz data set.}
\label{fig:brodatz_PR}
\end{figure}

Originally, precision/recall is defined only for two-classes problems. As there are more classes here, the curve is computed for each class, assuming the current class as the positive \textcolor{black}{prediction} and all the remaining as the negative \textcolor{black}{prediction}. Thus an average precision/recall curve is presented for discussion. Figure \ref{fig:brodatz_PR} shows the curves for Brodatz textures. A good method is supposed to have \textcolor{black}{a curve the closest possible} of a constant curve with precision always being 1. On average, this profile is satisfied by the proposed descriptors, once although it has not so high precision for low recall, it outperforms the other approaches for higher recall. Such behavior implies that the triangular descriptors do better in more complicated problems where more queries are required to obtain the correct result.

To quantify the performance on image retrieval a global measure may also be necessary and the most commonly used is the area under the curve of precision/\textcolor{black}{recall}. Table \ref{tab:AUC_brodatz} compares this measure for different methods on Brodatz textures. The proposed method has the largest computed area confirming its efficiency for image retrieval. Actually, the poor result for lower recalls is superimposed by the great precision when the number of queries increases.
\begin{table*}[!htpb]
	\centering
	\scriptsize
		\begin{tabular}{c c}
			\hline
                 Method        & Area Under Curve\\
                 \hline
									LBP          &  0.84\\
									GLCM         &  0.82\\
									Multifractal &  0.81\\                 
									Gabor        &  0.81\\ 									                                  
									Fourier      &  0.74\\
									Proposed     &  0.90\\
			\hline			
		\end{tabular}
	\caption{Areas under the curves of precision/recall for each compared descriptor on the Brodatz database.}
	\label{tab:AUC_brodatz}
\end{table*}

\begin{figure}[!htpb]
\centering
\includegraphics[width=.6\textwidth]{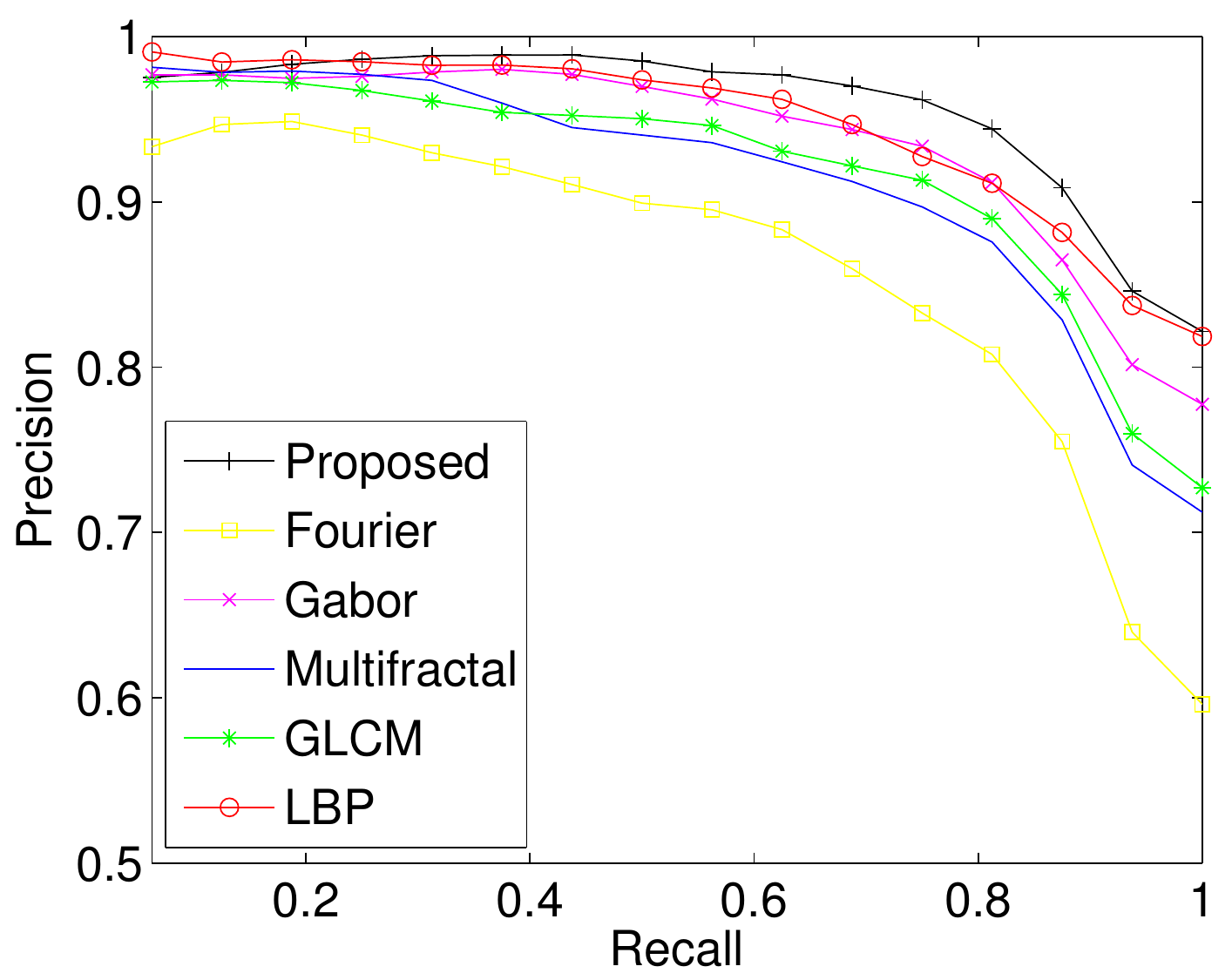}
\caption{Precision/Recall curves for Vistex data set.}
\label{fig:vistex_PR}
\end{figure}

Figure \ref{fig:vistex_PR} shows the precision/recall curves for Vistex database. Here, a similar phenomenon to that found in Figure \ref{fig:brodatz_PR} is repeated and the triangular descriptors have advantage for recalls close to 1. Now the precision for recalls close to 0 is \textcolor{black}{acceptable} yet it is not better than \textcolor{black}{state-of-the-art} methods such as LBP and Multifractal Spectrum. \textcolor{black}{In both cases, the proposed method behaves more regularly than the other approaches, making it a potential candidate to be applied in practical situations of image retrieval problems.}

The areas under curves are showed in Table \ref{tab:AUC_vistex}. Again, the triangular fractal descriptors have the largest area and have great potential to provide good outcomes in this kind of task.
\begin{table*}[!htpb]
	\centering
	\scriptsize
		\begin{tabular}{c c}
			\hline
                 Method        & Area Under Curve\\
                 \hline
									LBP          & 0.82\\
									GLCM         & 0.79\\
									Multifractal & 0.79\\                 
									Gabor        & 0.81\\ 									                                  
									Fourier      & 0.80\\
									Proposed     & 0.95\\
			\hline			
		\end{tabular}
	\caption{Areas under the curves of precision/recall for each compared descriptor on the Vistex database.}
	\label{tab:AUC_vistex}
\end{table*}

Generally speaking, in both data sets, the proposed method presents better results than their counterparts after a number of \textcolor{black}{retrieval attempts}. This is a consequence of combining different weights given by the exponents to each scale in the triangular descriptors. Although the empirical combination may not be enough for an initial guess, it ensures more robustness when more queries are requested.

\subsection{Image Classification}

In this section, the textures from the benchmark sets are classified. Since the actual classes are known previously, the accuracy of each approach can be measured and compared by statistical metrics.

\begin{figure}[!htpb]
\centering
\includegraphics[width=.6\textwidth]{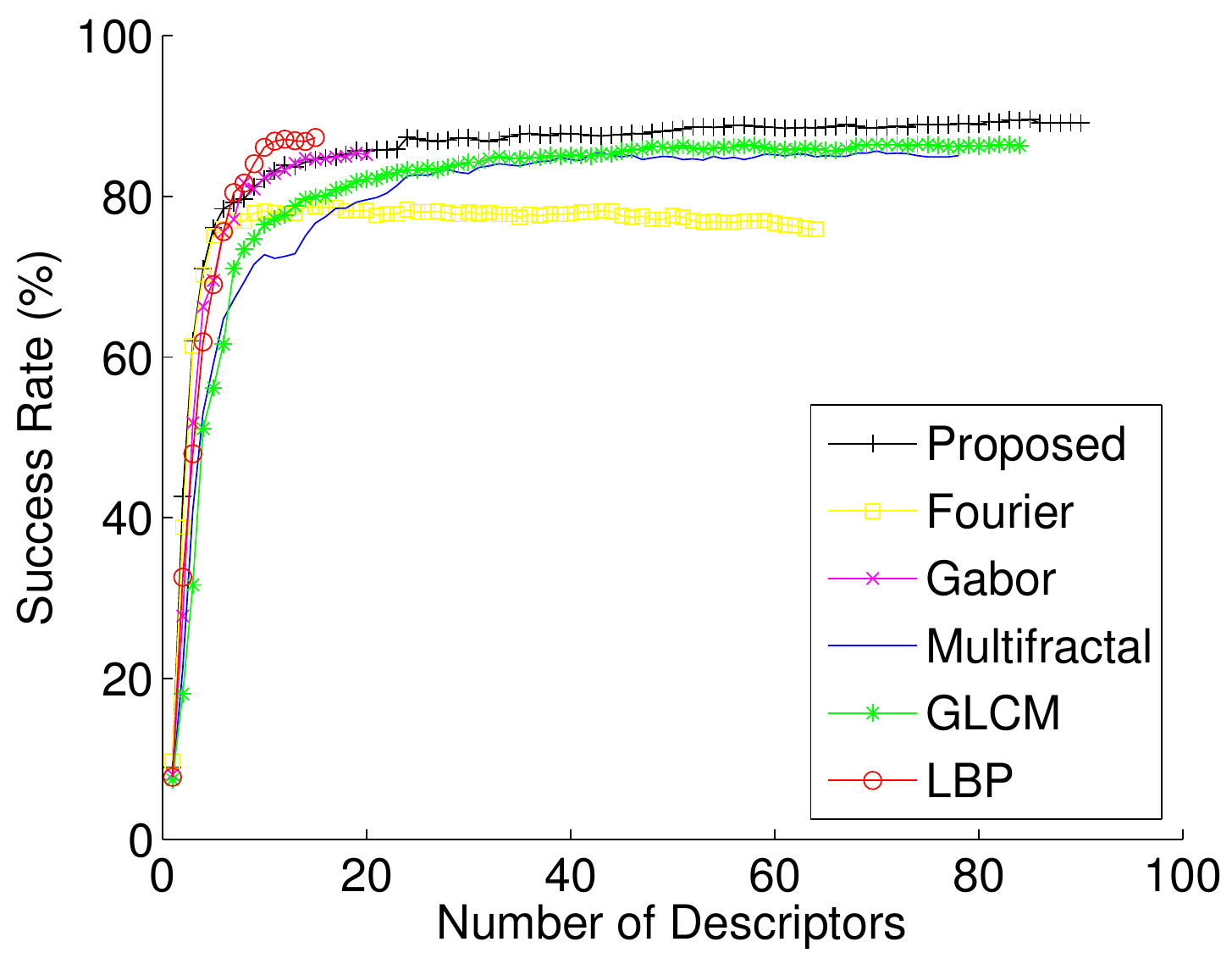}
\caption{Success rates in the classification of Brodatz textures according to the number of descriptors.}
\label{fig:correctness_brodatz}
\end{figure}

Figure \ref{fig:correctness_brodatz} shows in a plot the relation between the number of descriptors employed in the classification and the \textcolor{black}{rate of images correctly classified (success rate)} obtained in Brodatz data set. Even though the state-of-the-art LBP method is the best when using a few descriptors, the triangular fractal descriptors ensures a more accurate classification using more than 34 features, which is a reasonable number given that the error is kept within an acceptable range.

Table \ref{tab:CR_brodatz} shows the \textcolor{black}{success} rates, \textcolor{black}{the} respective errors and number of descriptors for each compared texture descriptors. The proposed descriptors achieved an advantage of more than 2\% over the second best approach (LBP). Such gain is relevant given the size and number of classes in this database, which makes such rates close to the limit of best possible results.  Moreover, if on the one hand more descriptors are necessary, on the other the error in the cross validation is lower, showing that no dimensionality curse is detected.
\begin{table*}[!htpb]
	\centering
	\scriptsize
		\begin{tabular}{c c c}
			\hline
                 Method        & \textcolor{black}{Success} rate & Number of descriptors\\
                 \hline
									LBP          &  87.33$\pm$0.02  & 15\\
									GLCM         &  86.48$\pm$0.02  & 70\\
									Multifractal &  85.64$\pm$0.03  & 70\\                 
									Gabor        &  85.42$\pm$0.02  & 19\\ 																Fourier      &  78.71$\pm$0.03  & 15\\
									Proposed     &  89.58$\pm$0.01  & 85\\
			\hline			
		\end{tabular}
	\caption{\textcolor{black}{Success} rates (with respective errors) and number of descriptors for each compared method on Brodatz data.}
	\label{tab:CR_brodatz}
\end{table*}

\begin{figure}[!htpb]
					 \centering
           \mbox{\subfigure{\includegraphics[width=0.5\textwidth]{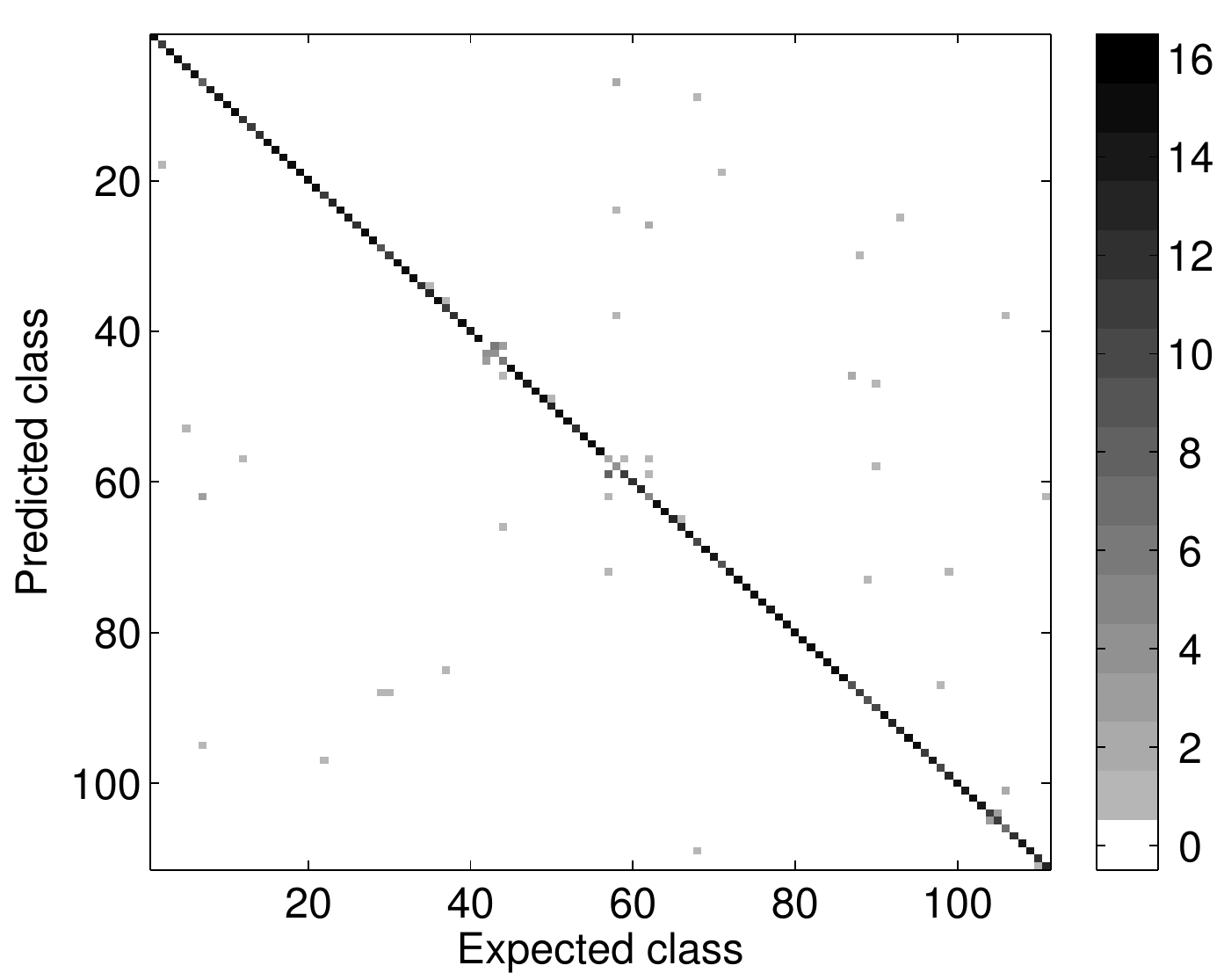}}
           			 \subfigure{\includegraphics[width=0.5\textwidth]{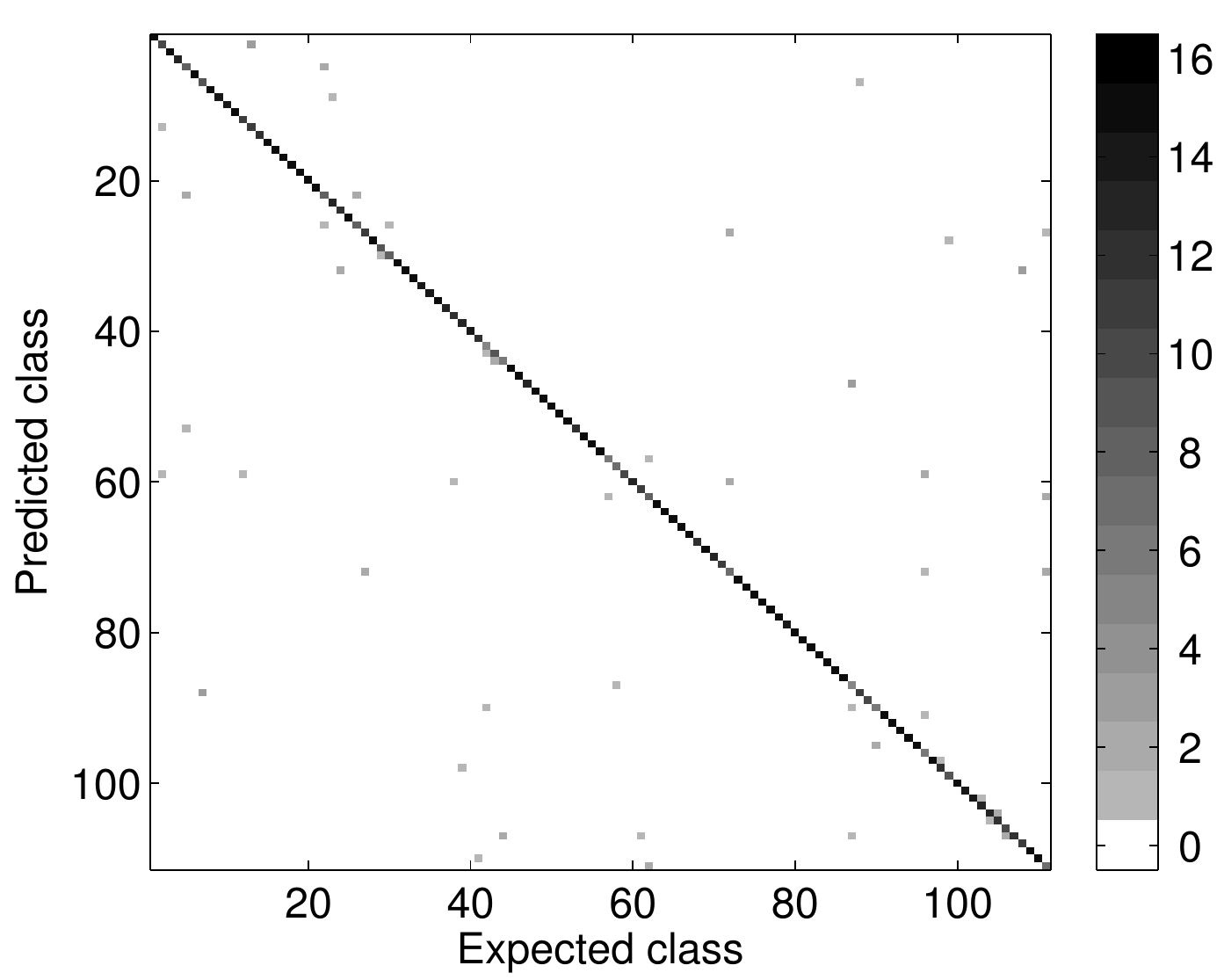}}}
					 \mbox{\subfigure{\includegraphics[width=0.5\textwidth]{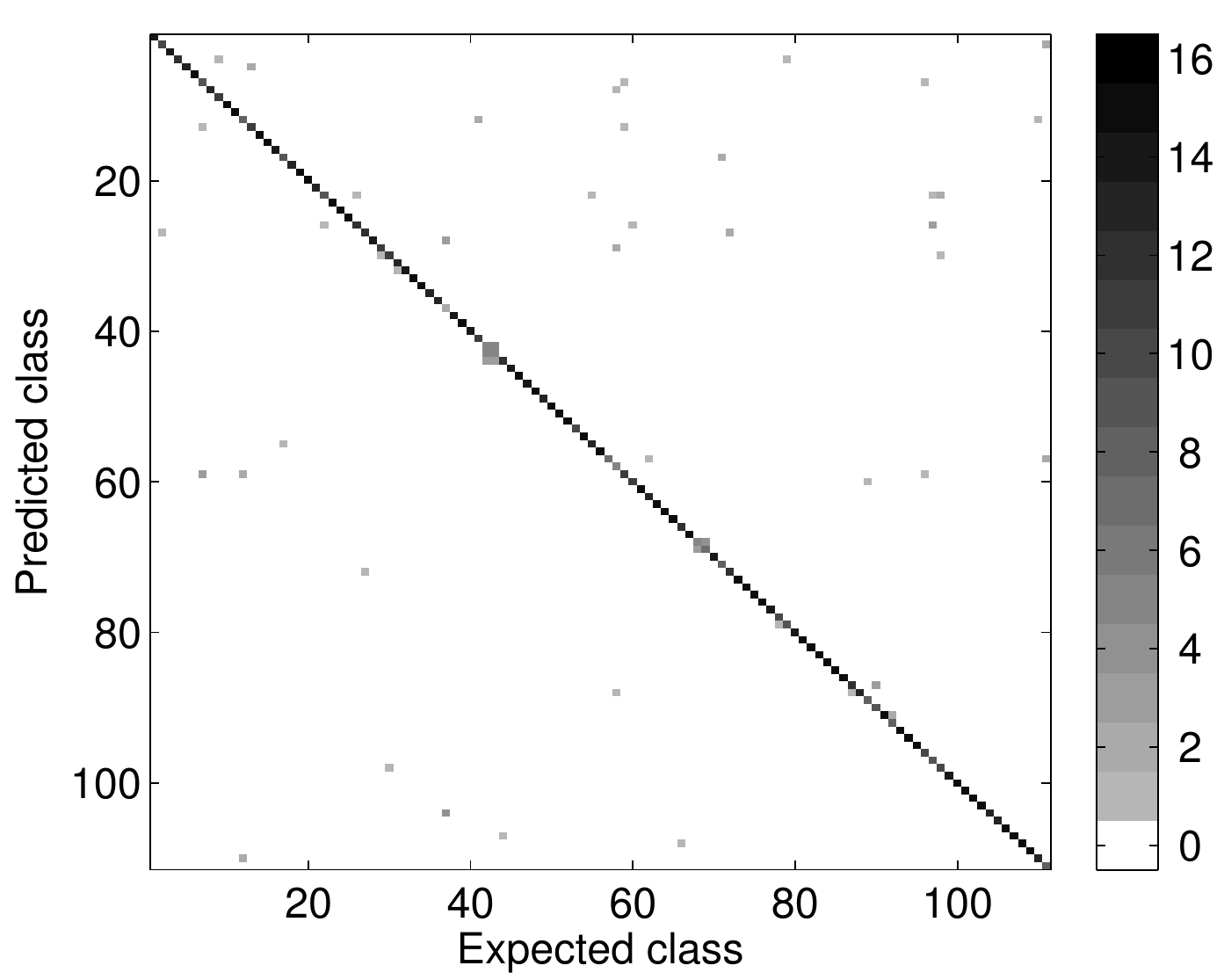}}
                 \subfigure{\includegraphics[width=0.5\textwidth]{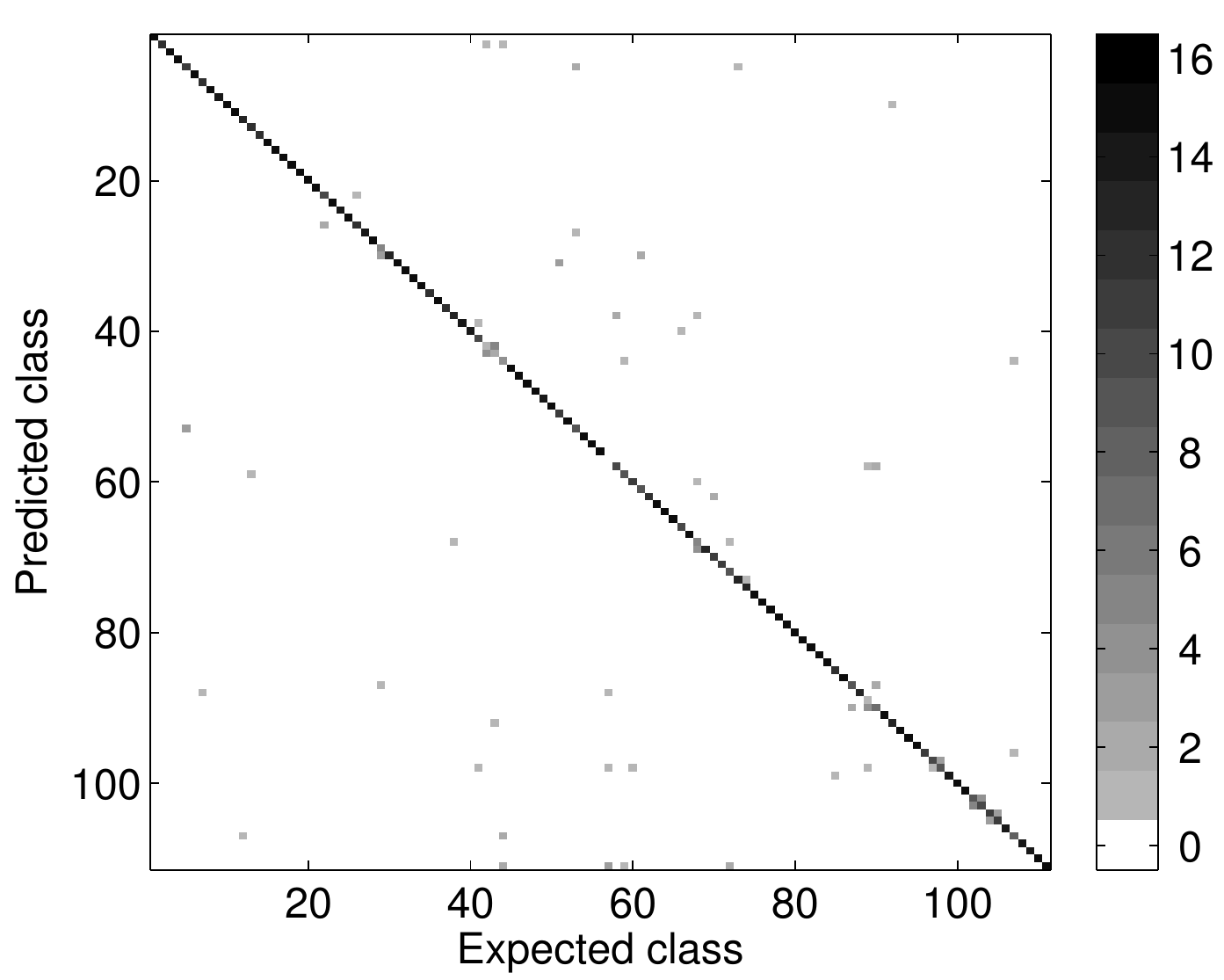}}}
           \mbox{\subfigure{\includegraphics[width=0.5\textwidth]{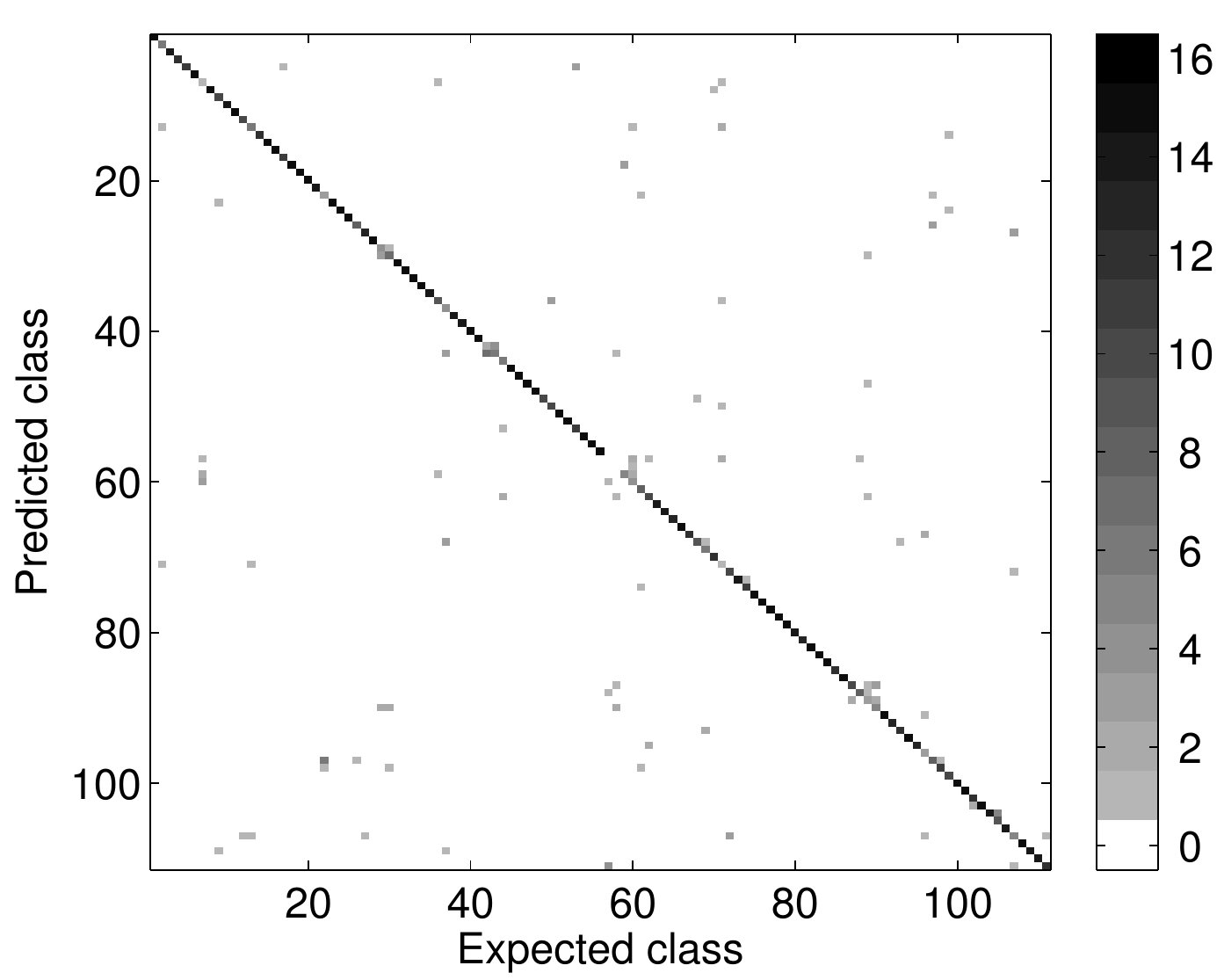}}
					 	     \subfigure{\includegraphics[width=0.5\textwidth]{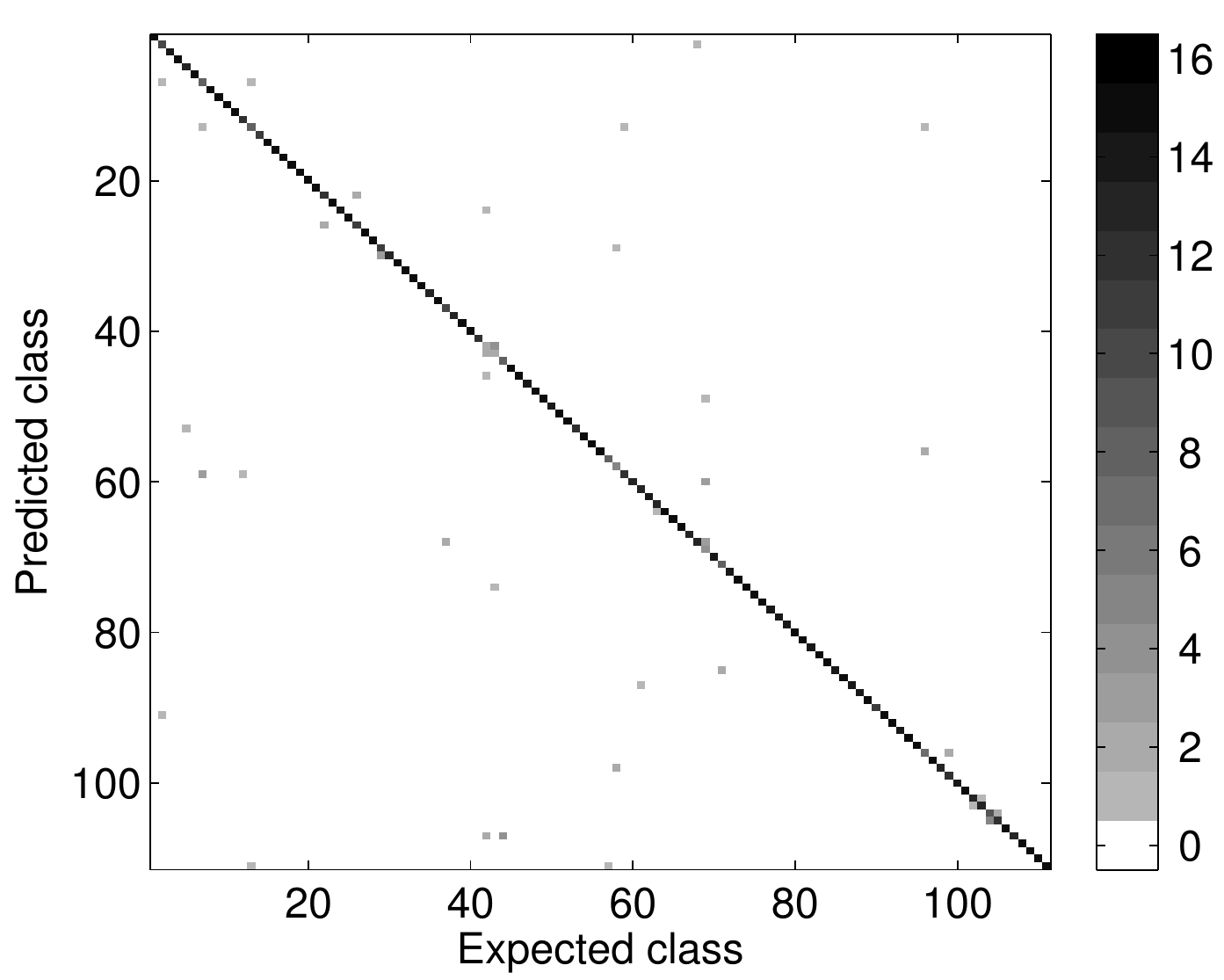}}}								 					 	     
           \caption{Confusion marices for the methods on Brodatz data set. (a) LBP. (b) GLCM. (c) Multifractal. (d) Gabor. (e) Fourier. (f) Proposed method.}
           \label{fig:CM_brodatz}                                  
   \end{figure}

Figure \ref{fig:CM_brodatz} shows the confusion matrices for the respective descriptors. Such matrices expressing the number of images assigned to each class is represented in a grayscale image. In these diagrams, an ideal method should present a black diagonal with a clear white background. Any light point in the diagonal and gray point outside corresponds to errors in the classification. The primary goal of this type of expression is to show the behavior of the descriptors in each class. For instance, both triangular and LBP methods come off badly in the classes 42/43. Those classes do not have enough texture information, but they are more suitable for a shape-based analysis. On the other hand, they completely differ in classes like 69 and 62. Images from class 69 exhibit different macro-patterns that impair the fractal-based method, but not local approaches such as LBP. The opposite situation arises with class 62, where there are local illumination changes but the multiscale patterns are preserved.

\begin{figure}[!htpb]
\centering
\includegraphics[width=.6\textwidth]{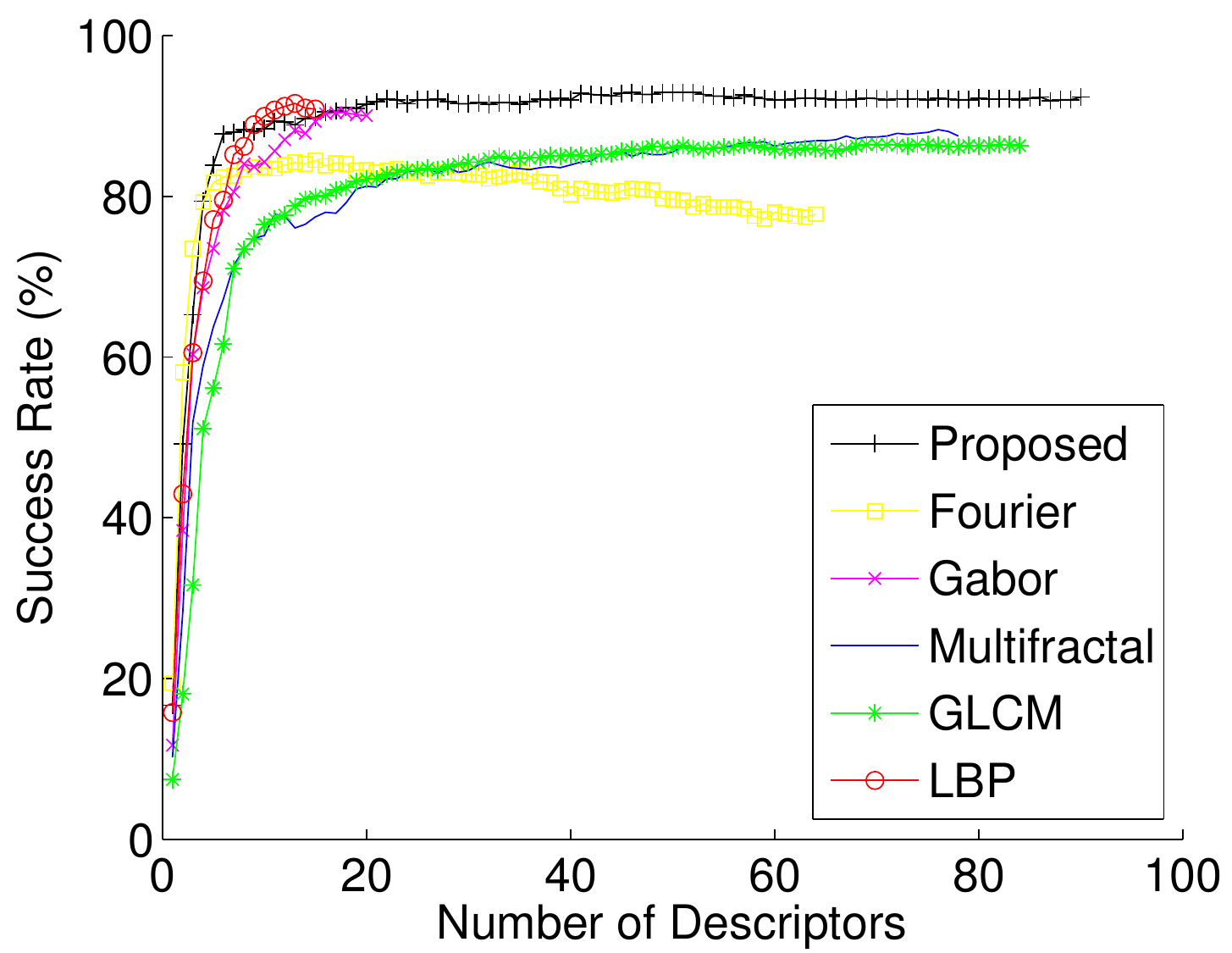}
\caption{Success rates in the classification of Vistex textures according to the number of descriptors.}
\label{fig:correctness_vistex}
\end{figure}

Figure \ref{fig:correctness_vistex} shows the relation \textcolor{black}{success} rate/number of descriptors for Vistex. The general \textcolor{black}{shape} of each curve is similar to those in Figure \ref{fig:correctness_brodatz}, confirming the reproducibility of the experiment. The highest success rates are caused by the smaller number of classes, even though each Vistex image is more challenging than Brodatz, as it has a higher level of variances intra-classes.

Table \ref{tab:CR_vistex} shows the success rates for Vistex. It is worth to notice that a number of triangular descriptors smaller than in Brodatz case are necessary to obtain the best result. This is also a consequence of the reduced number of classes, which makes the problem statistically simpler.
\begin{table*}[!htpb]
	\centering
	\scriptsize
		\begin{tabular}{c c c}
			\hline
                 Method        & \textcolor{black}{Success} rate (\%) & Number of descriptors\\
                 \hline
									LBP          & 91.55$\pm$0.03 & 13\\
									GLCM         & 88.21$\pm$0.03 & 70\\
									Multifractal & 88.31$\pm$0.03 & 76\\                 						
									Gabor        & 90.39$\pm$0.01 & 17\\ 								                                  	
									Fourier      & 84.49$\pm$0.02 & 15\\
									Proposed     & 92.94$\pm$0.01 & 50\\
			\hline			
		\end{tabular}
	\caption{\textcolor{black}{Success} rates (with respective errors) and number of descriptors for each compared method on Vistex data.}
	\label{tab:CR_vistex}
\end{table*}

\begin{figure}[!htpb]
					 \centering
           \mbox{\subfigure{\includegraphics[width=0.5\textwidth]{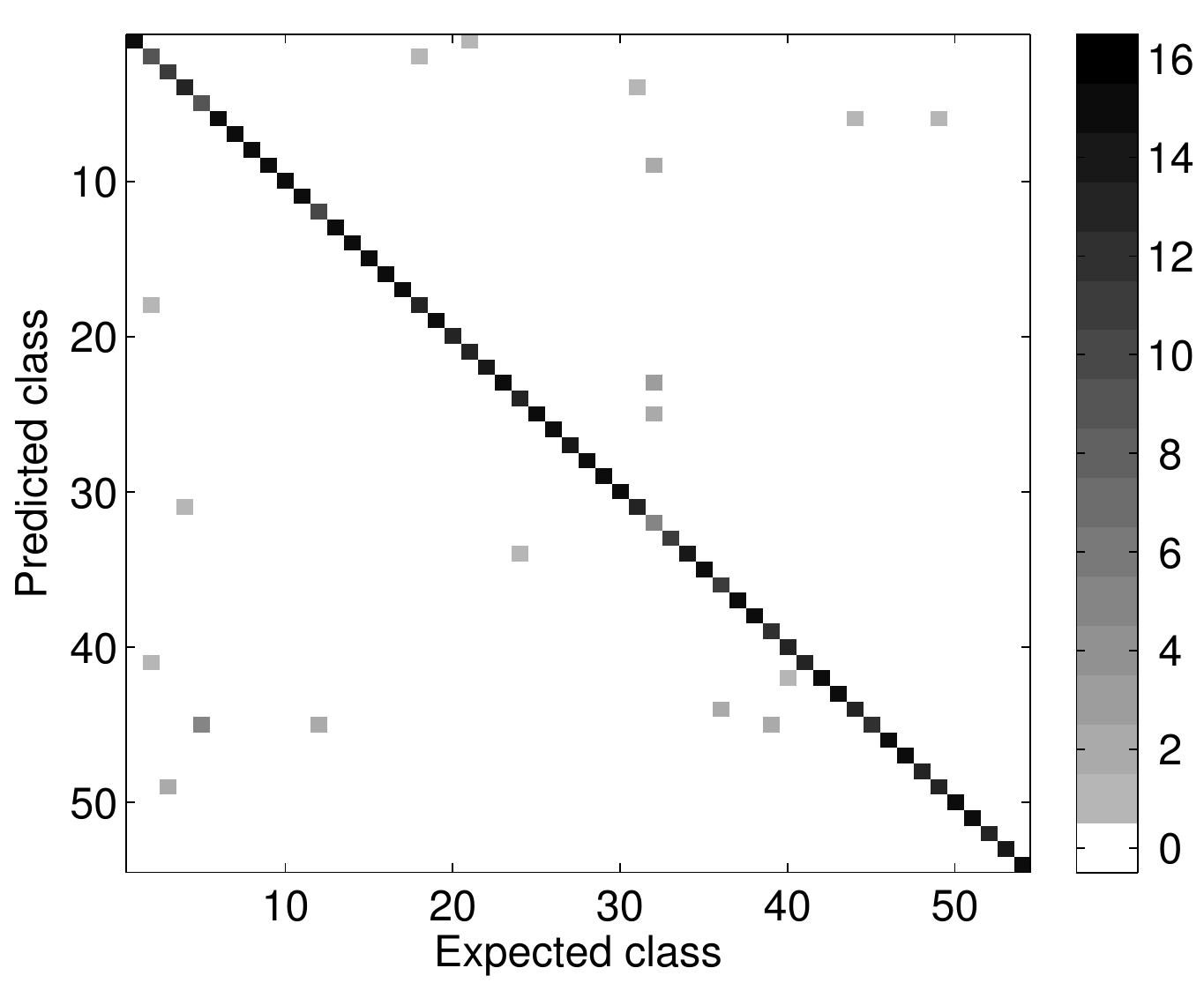}}
           			 \subfigure{\includegraphics[width=0.5\textwidth]{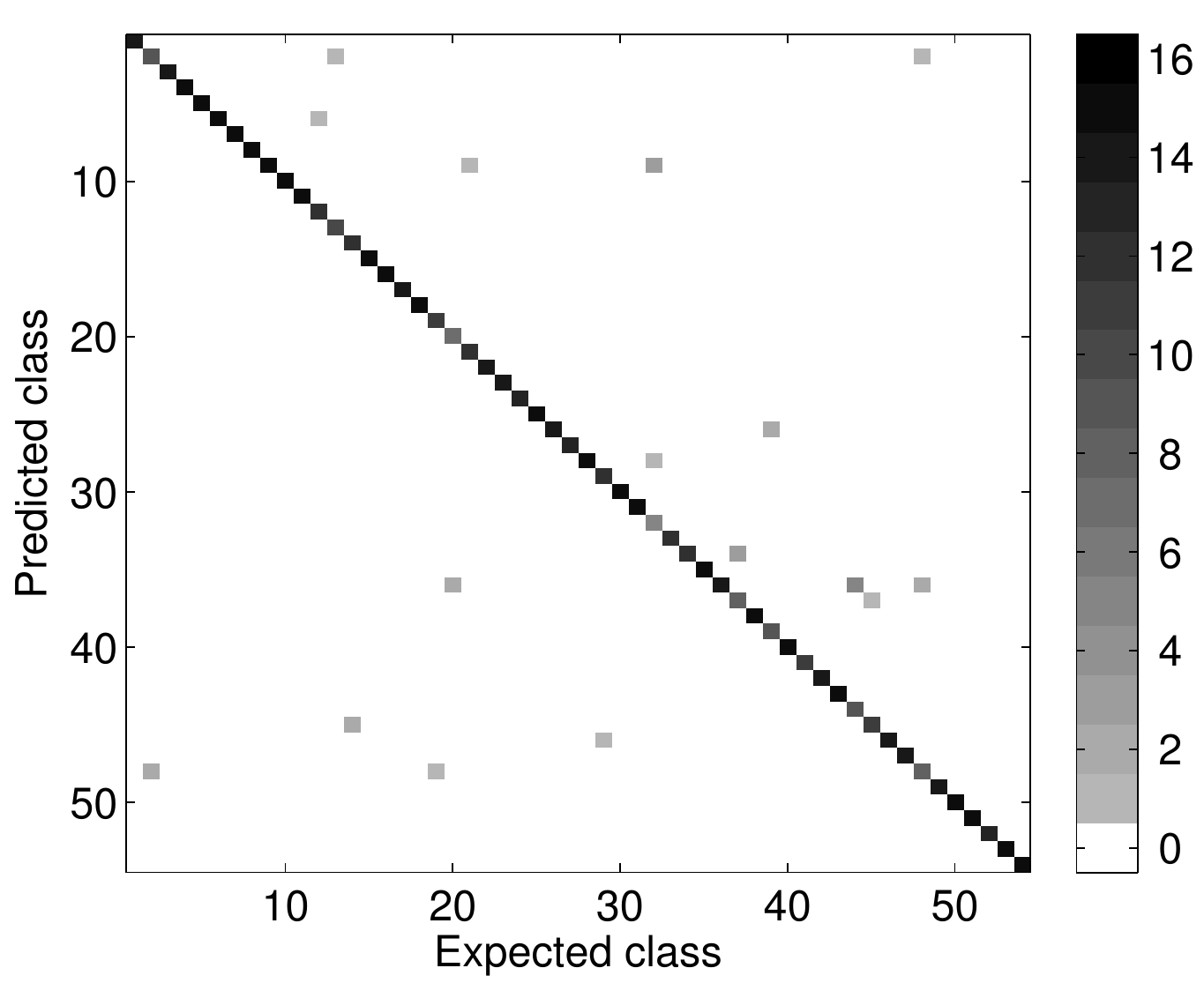}}}
					 \mbox{\subfigure{\includegraphics[width=0.5\textwidth]{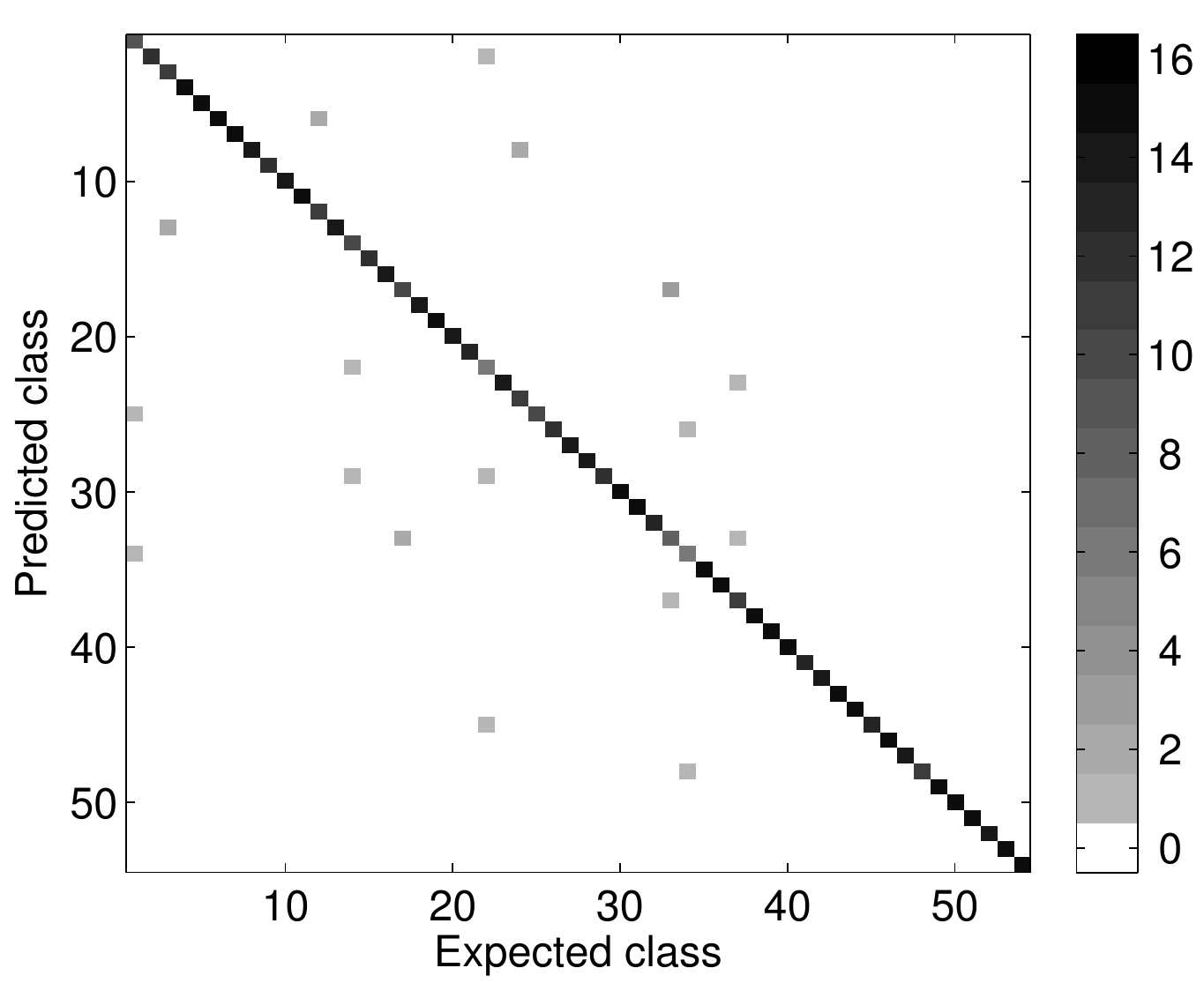}}
                 \subfigure{\includegraphics[width=0.5\textwidth]{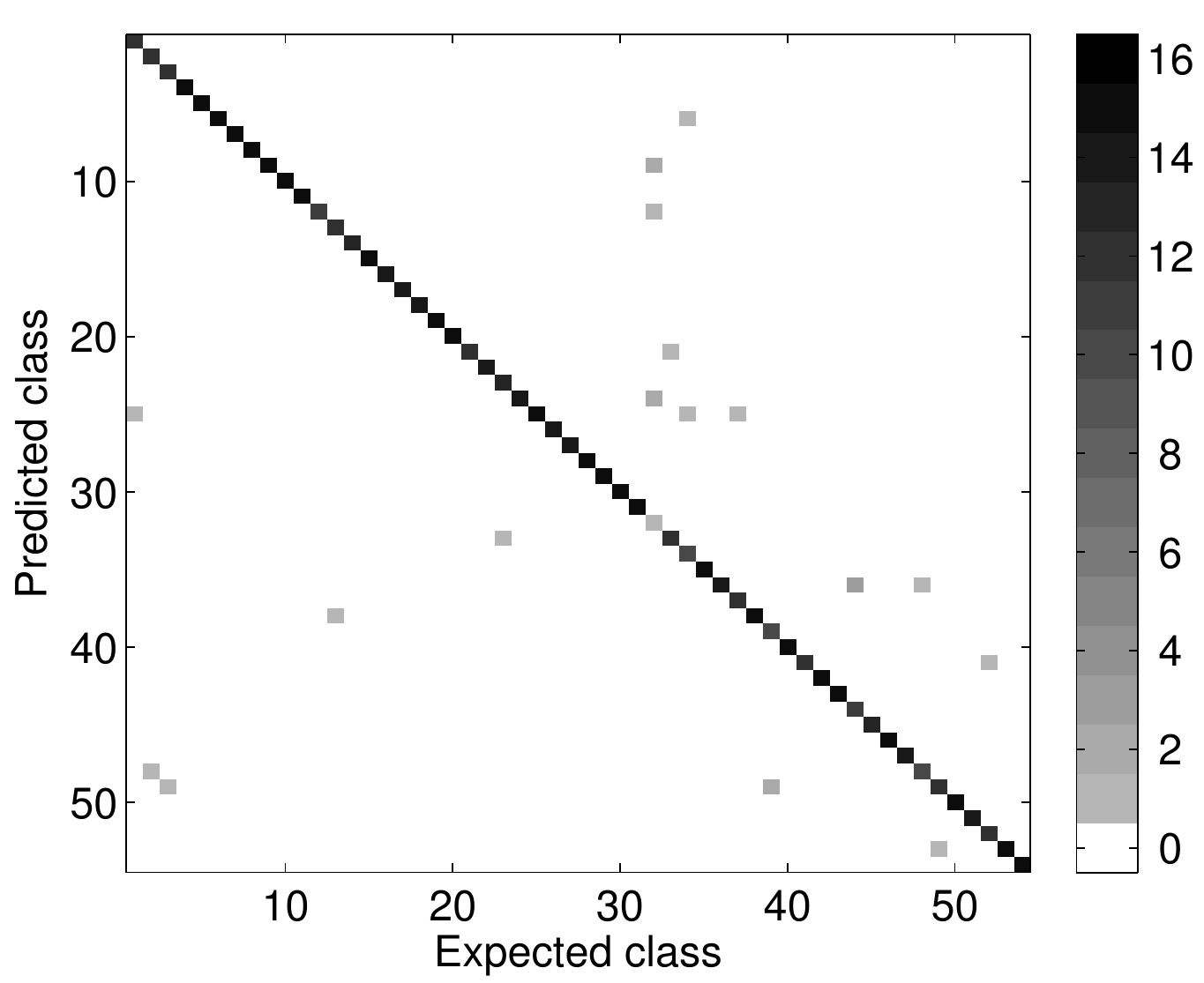}}}
           \mbox{\subfigure{\includegraphics[width=0.5\textwidth]{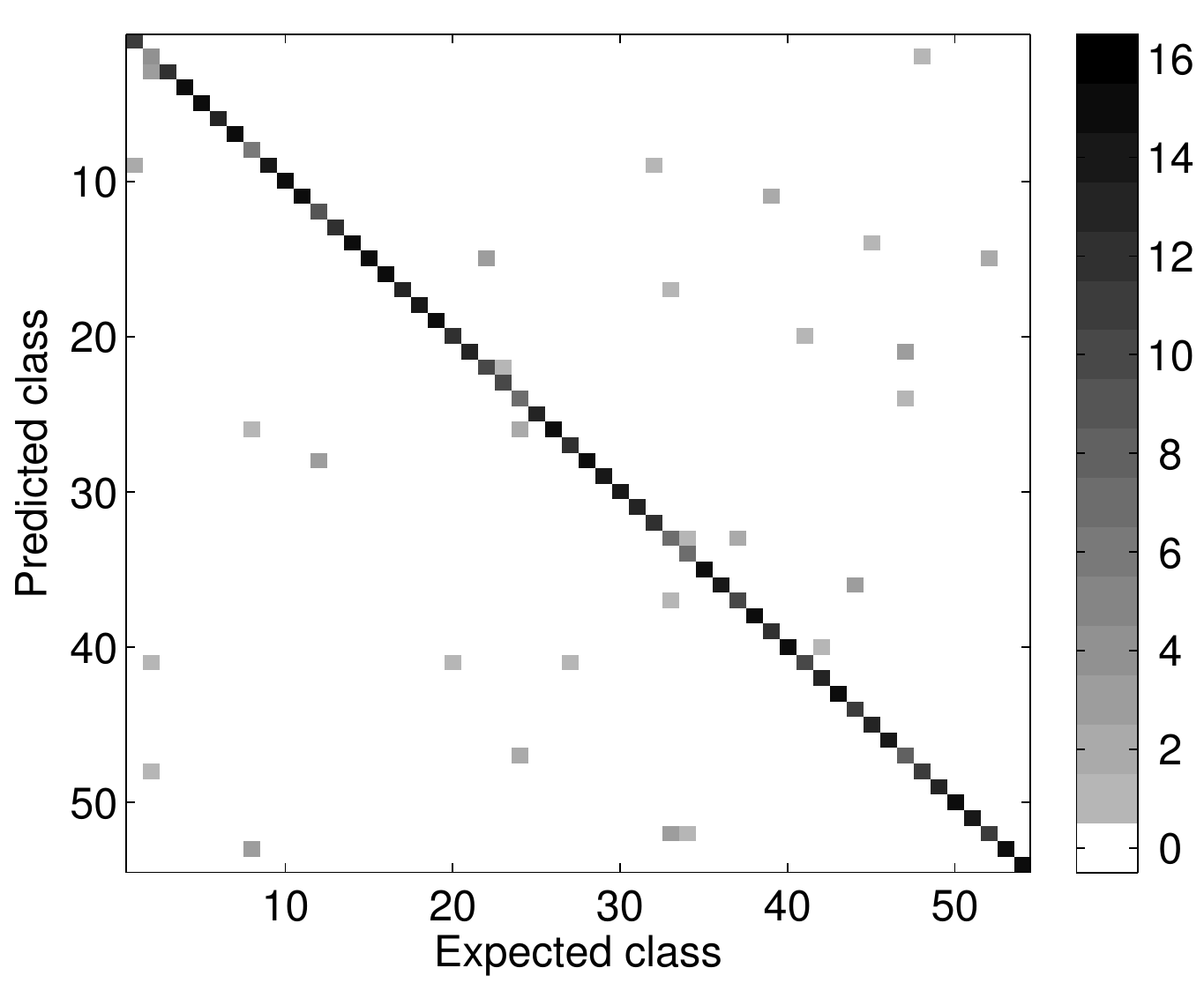}}
					 	     \subfigure{\includegraphics[width=0.5\textwidth]{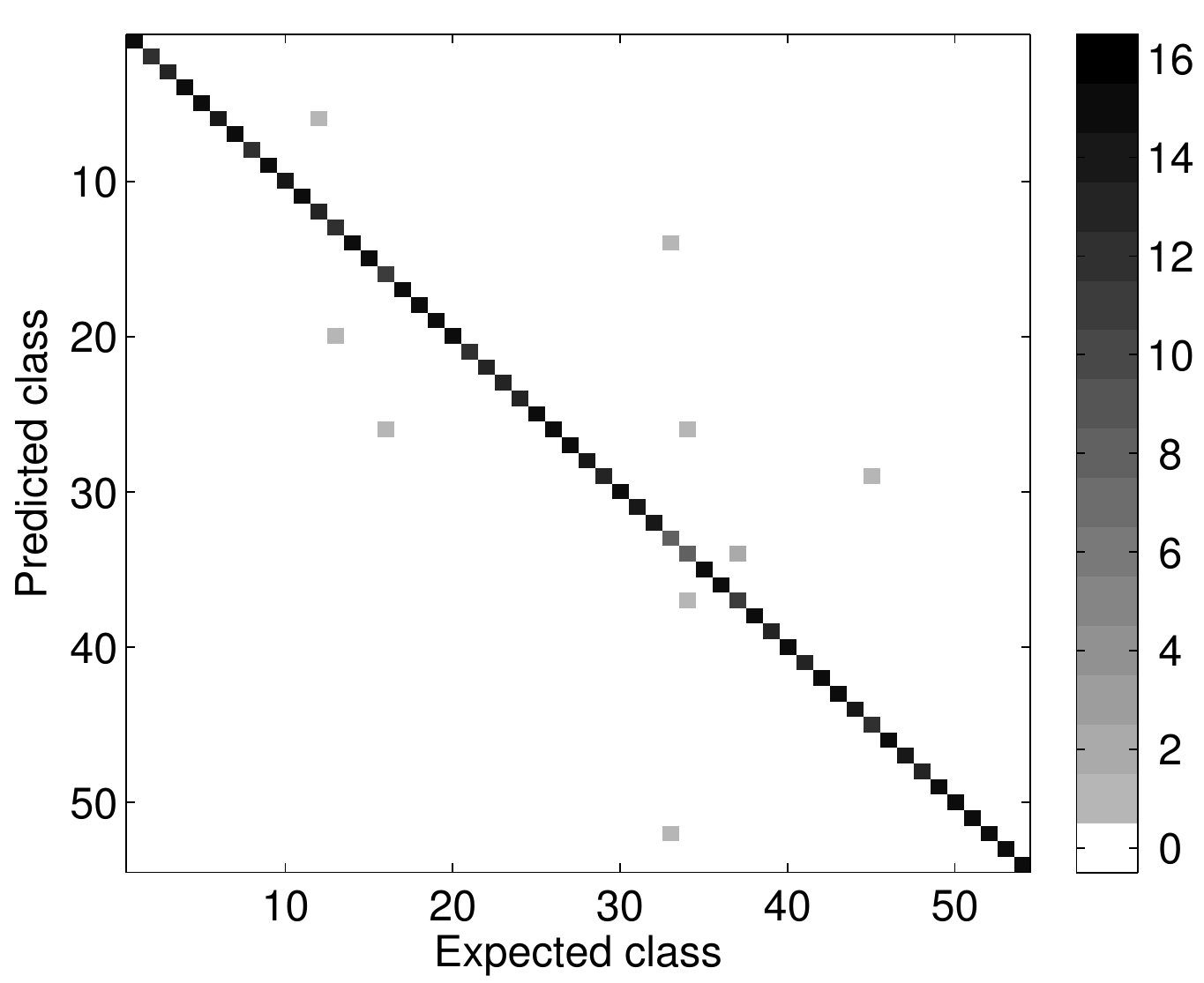}}}								 					 	     
           \caption{Confusion marices for the methods on Vistex data set. (a) LBP. (b) GLCM. (c) Multifractal. (d) Gabor. (e) Fourier. (f) Proposed method.}
           \label{fig:CM_vistex}                                  
   \end{figure}

Figure \ref{fig:CM_vistex} shows the confusion matrices for Vistex database. The best three descriptors (Gabor, LBP and Triangular) have matrices visually similar, but the behavior in each class \textcolor{black}{differs} among the approaches. For example, the class 34 has different global patterns that can be confused by methods like Gabor and Triangular, while the class 5 has large homogeneous regions that \textcolor{black}{impair} the performance of local methods like LBP.

\subsection{Invariances}

Another important test to assess the robustness of any image descriptor is to verify its invariance to changes like rotation and addition of noise.

\begin{figure}[!htpb]
	\centering
           \mbox{\subfigure{\includegraphics[width=0.1\textwidth]{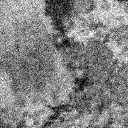}}
				\subfigure{\includegraphics[width=0.1\textwidth]{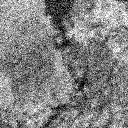}}
				\subfigure{\includegraphics[width=0.1\textwidth]{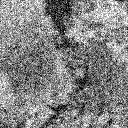}}
				\subfigure{\includegraphics[width=0.1\textwidth]{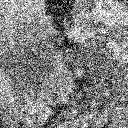}}
				\subfigure{\includegraphics[width=0.1\textwidth]{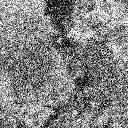}}}		
	       \caption{A texture with different levels of Gaussian noise. (a) 0.01. (b) 0.02. (c) 0.03. (d) 0.04. (e) 0.05.}
           \label{fig:noise}                                  
   \end{figure}

The first evaluated invariance is to \textcolor{black}{noise} following Gaussian distribution, a situation commonly found in practice. The levels of noise (ratio of affected pixels) ranges between 0.01 and 0.05. Figure \ref{fig:noise} shows examples of such images affected by the different \textcolor{black}{levels of noise in our tests}.

Table \ref{tab:noise} presents results comparing with other descriptors in Brodatz data set. A similar behavior is expected for other textures. Although Gabor obtained the best results, mainly for more severe noises, the proposed method does better than the state-of-the-art LBP method and for a small amount of noise it has a performance similar to Gabor.
\begin{table*}[!htpb]
	\centering
	\scriptsize
		\begin{tabular}{c c c c c c}
			\hline
                   Method        & \multicolumn{5}{c}{Noise ratio}\\
                   \hline
                                & 0.01  & 0.02  & 0.03  & 0.04  & 0.05\\
                   \hline
					LBP          & 71.23 & 69.88 & 65.14 & 62.39 & 59.79\\
					GLCM         & 80.46 & 78.10 & 78.21 & 78.04 & 77.65\\
					Multifractal & 76.58 & 71.00 & 66.10 & 65.48 & 63.79\\                 								Gabor        & 83.67 & 83.73 & 83.22 & 82.77 & 82.38\\ 												Fourier      & 78.99 & 78.71 & 78.43 & 78.15 & 77.59\\
					Proposed     & 83.33 & 81.64 & 79.05 & 77.36 & 76.13\\
			\hline			
		\end{tabular}
	\caption{\textcolor{black}{Success} rates for different levels of Gaussian noise applied to Brodatz data.}
	\label{tab:noise}
\end{table*}

\begin{figure}[!htpb]
\centering
\includegraphics[width=.2\textwidth]{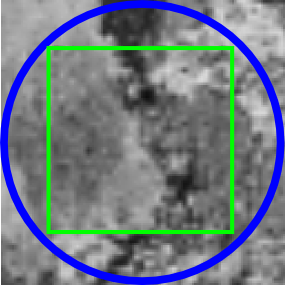}
\caption{Process to obtain a central region of the rotated texture. The region of interest selected from the texture is that inside the green square.}
\label{fig:rot_process}
\end{figure}

A second invariance experiment evaluates textures rotated by predefined angles. To ensure the same \textcolor{black}{dimension} for all textures, a procedure to extract only the central region of each image is accomplished, as illustrated in Figure \ref{fig:rot_process}. In this way, if the original texture image has dimensions $d \times d$, \textcolor{black}{the} extracted region after the rotation is the central part with \textcolor{black}{dimension} $d/\sqrt{2} \times d/\sqrt{2}$.

\begin{figure}[!htpb]
	\centering
           \mbox{\subfigure{\includegraphics[width=0.1\textwidth]{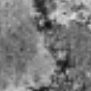}}
				\subfigure{\includegraphics[width=0.1\textwidth]{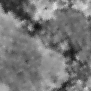}}
				\subfigure{\includegraphics[width=0.1\textwidth]{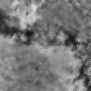}}
				\subfigure{\includegraphics[width=0.1\textwidth]{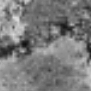}}
				\subfigure{\includegraphics[width=0.1\textwidth]{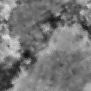}}
				\subfigure{\includegraphics[width=0.1\textwidth]{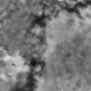}}
				\subfigure{\includegraphics[width=0.1\textwidth]{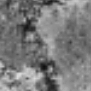}}}		
	       \caption{A texture rotated by different angles. From left to right, 0, 30, 60, 90, 120, 150 and 180 degrees.}
           \label{fig:rot}                                  
   \end{figure}

Figure \ref{fig:rot} shows the example of a texture rotated by the angles considered in this experiment, that is, 0, 30, 60 90, 120, 150 and 180 degrees.

Table \ref{tab:rotation} presents the correctness rates for the classification of Brodatz database after the rotation of the textures. The lowest rates are caused by the use of a cropped region of the original texture, leading to an obvious loss of information. The proposed approach outperforms the \textcolor{black}{other} methods with higher success rates and a low variability in the results when the angle is changed. The maximum difference is only a little over 1\%, while LBP, for example, shows a difference of almost 10\% in the result when the textures are rotated by 150 and 180 degrees. 
\begin{table*}[!htpb]
	\centering
	\scriptsize
		\begin{tabular}{c c c c c c c c c}
			\hline
                   Method        & \multicolumn{7}{c}{Rotation angle (degrees)}\\
                   \hline
                                & 0  & 30  & 60  & 90  & 120  & 150  & 180\\
                   \hline
					LBP          & 81.98 & 73.70 & 72.86 & 81.98 & 73.98 & 72.75 & 82.15\\
					GLCM         & 77.70 & 75.28 & 75.50 & 77.93 & 75.50 & 76.23 & 77.53\\
					Multifractal & 76.63 & 76.24 & 73.99 & 77.25 & 76.12 & 76.18 & 76.63\\                					Gabor        & 80.13 & 80.41 & 79.06 & 80.01 & 80.13 & 79.11 & 80.29\\ 								Fourier      & 70.32 & 72.07 & 71.56 & 69.98 & 72.07 & 71.78 & 70.21\\
					Proposed     & 83.33 & 84.69 & 84.18 & 83.67 & 84.58 & 83.84 & 83.56\\
			\hline			
		\end{tabular}
	\caption{\textcolor{black}{Success} rates for different angles of rotation applied to Brodatz data.}
	\label{tab:rotation}
\end{table*}

\color{black}

\subsubsection{Discussion}

The proposed method differs from other approaches to texture analysis in the literature in that the physical process underlying the generation of the image is taken into account to a high extent, as we demonstrated its relation with fractional Brownian processes. These processes are known to be widely present in nature and its association with the way that our brain perceives materials ans scenarios around us is well established. Parallel to this, the triangular prisms also encompass a geometrical analysis similar to ``walking-dividers'', intimately related to the primitive idea of fractals and also commonly observed in natural structures. In itself, such complementary and dual viewpoint of fractality is an important contribution and novelty with regards to other fractal approaches such as multifractals \cite{XJF09}, Bouligand-Minkowski \cite{BCB09}, and others.

The consideration of the physical process also showed to be more advantageous than simply quantifying relations between neighbor pixels without accounting for the semantic involved in the represented object, as in LBP and GLCM descriptors. Even though the interpretation of these approaches can be considered more straightforward, the lack of a more realistic model makes them insufficient in more complex cases with larger and more heterogeneous databases as those presented here.

Finally, we also compare to other methods where the image is observed beyond the simple pixel values, such as in Fourier and Gabor descriptors. These descriptors, however, are ultimately based on linear filtering and are not the ideal solution to explain the non-linearities present in many natural images, mostly caused by the chaotic behavior associated to fBm processes.

Generally speaking, the great performance of triangular fractal descriptors in the presented results confirm what is expected from the fractal analysis of natural texture images. Indeed, fractal geometry constitutes a reliable tool to model those structures, which cannot be well-described in an Euclidean framework. This analysis is especially complete here as two fractal viewpoints (statistical and geometrical) are merged to provide an even more accurate description of the image. Furthermore, fractal descriptors approach enhances the conventional fractal geometry analysis by extracting relevant information of complexity under different scales, by the multiscale analysis implicit when the dimension is estimated for different values of $\epsilon$. \textcolor{black}{The combination of a solid modeling with a multiscale context also explains the robustness to noise and rotation, as noise is more active in local scales and deformations like the image rotation do not alter model parameters at a substantial level.} In this way, these descriptors are capable of providing features representing precise and rich measures of the spatial structure of complex textures like those in benchmark data, as used here, as well as in practical application involving the identification and discrimination of intricate patterns in objects and scenarios represented in a digital image.

\color{black}
   
\section{Conclusion}

This work proposed a novel gray-level texture descriptor based on a particular method of fractal dimension estimation, named triangular prism method. The descriptors were obtained by combining values of area of a triangular tessellation of the texture, using different steps for the tessellation grid as well as different values for an exponential parameter used as a weight for the area within each cell in the grid.

\textcolor{black}{Our study demonstrated that the triangular prisms are capable of extracting fractal characteristics of the image under two complementary perspectives: geometrical (walking-dividers) and statistical (fractional Brownian motion).}

These descriptors were compared to classical and state-of-the-art texture descriptors in a task of classification of two well-known benchmark texture datasets, e.g., Brodatz and Vistex.

The classification results illustrated that the novel approach is a valuable descriptor, achieving remarkable results in image classification and retrieval tasks and suggesting its application to a wide range of problems comprising tasks of pattern recognition in digital images.

\section*{Acknowledgements}
O. M. Bruno gratefully acknowledges the financial support of CNPq (National Council for Scientific and Technological Development, Brazil) (Grant \#307797/2014-7 and Grant \#484312/2013-8) and FAPESP (The State of S\~ao Paulo Research Foundation) (Grant \# 14/08026-1).
J. B. Florindo gratefully acknowledges the financial support of FAPESP Proc. 2013/22205-3 and 2012/19143-3.

%\bibliographystyle{unsrt}
%\bibliographystyle{myplain}
%\bibliography{Triangular}

\end{document}